\documentclass[conference]{IEEEtran}

\IEEEoverridecommandlockouts

\usepackage{amsmath,amssymb,amsfonts}   
\usepackage{graphicx}                   
\usepackage{textcomp}                   
\usepackage{xcolor}                     
\usepackage{longtable}                  
\usepackage{array} 
\usepackage{multirow}                   
\usepackage{soul}                       

\usepackage{algorithm2e}                
\usepackage{algpseudocode}              

\usepackage[colorlinks=true,linkcolor=blue,citecolor=blue,urlcolor=blue]{hyperref}                 
\usepackage[most]{tcolorbox}
\usepackage{graphbox}
\usepackage{tabularx}
\usepackage{booktabs}
\newcolumntype{P}[1]{>{\centering\arraybackslash}p{#1}} 
\newcolumntype{L}{>{\raggedright\arraybackslash}X}
\usepackage{tikz}
\usetikzlibrary{shapes,arrows.meta,positioning}

\usepackage{bm}                         

\def\BibTeX{{\rm B\kern-.05em{\sc i\kern-.025em b}\kern-.08em T\kern-.1667em\lower.7ex\hbox{E}\kern-.125emX}}

\DeclareUnicodeCharacter{2212}{\ensuremath{-}}
\newcommand{\mycopyrightnotice}{}

\makeatletter
\def\ps@IEEEtitlepagestyle{%
  \def\@oddfoot{\mycopyrightnotice}%
  \def\@oddhead{\hbox{}\@IEEEheaderstyle\leftmark\hfil\thepage}\relax
  \def\@evenhead{\@IEEEheaderstyle\thepage\hfil\leftmark\hbox{}}\relax
  \def\@evenfoot{}%
}

\begin{document}

\title{Generative Deep Learning for Computational Destaining and Restaining of Unregistered Digital Pathology Images
\\
{\footnotesize \textsuperscript{}}
\thanks{}
}

\author{
    \IEEEauthorblockN{1\textsuperscript{st} Aarushi Kulkarni}
    \IEEEauthorblockA{\textit{Department of Computer Science} \\
    \textit{University of California}\\
    Irvine, CA, USA \\
    aarushk2@uci.edu}
    \and
    \IEEEauthorblockN{2\textsuperscript{nd} Alarice Lowe}\
    \IEEEauthorblockA{\textit{Department of Pathology} \\
    \textit{Stanford University}\\
    Stanford, CA, USA \\
    aclowe@stanford.edu}
    \and
    \IEEEauthorblockN{3\textsuperscript{rd} Pratik Shah*\thanks{*Corresponding author: Dr. Pratik Shah Ph.D. (pratik.shah@uci.edu)}}
    \IEEEauthorblockA{\textit{Department of Pathology and Laboratory Medicine} \\
    \textit{Biomedical Engineering} \\
    \textit{Electrical Engineering and Computer Science} \\
    \textit{University of California}\\
    Irvine, CA, USA \\
    pratik.shah@uci.edu}
}

\maketitle

\begin{abstract}
Conditional generative adversarial networks (cGANs) have enabled high-fidelity computational staining and destaining of hematoxylin and eosin (H\&E) in digital pathology whole-slide images (WSI). However, their ability to generalize to out-of-distribution WSI across institutions without retraining remains insufficiently characterized, limiting scalable clinical translation. Previously developed cGAN models trained on 102 registered prostate core biopsy WSIs from Brigham and Women's Hospital were evaluated on 82 spatially unregistered WSIs acquired at Stanford University. To mitigate domain shift without retraining, a preprocessing pipeline consisting of histogram-based stain normalization for H\&E-stained WSIs and channel-wise intensity calibration for unstained WSIs was developed. Because image registration was intentionally omitted for real-world deployment conditions, the reported quantitative results are conservative lower bounds reflecting both model performance and limited spatial alignment. Under these conditions, virtual destaining achieved a Pearson correlation coefficient (PCC) of 0.854, structural similarity index measure (SSIM) of 0.699, and peak signal-to-noise ratio (PSNR) of 18.41 dB. H\&E restaining from computationally destained outputs outperformed direct staining from ground-truth unstained inputs across all metrics (PCC: 0.798 vs. 0.715; SSIM: 0.756 vs. 0.718; PSNR: 20.08 vs. 18.51 dB), suggesting that preprocessing quality may be more limiting than model capacity. Qualitative pathological review indicated preservation of benign glandular structures while showing that malignant glands were often rendered with vessel-like morphologies. These findings support the feasibility of applying cGAN-based computational H\&E staining and destaining generative models to external WSI datasets using preprocessing-based adaptation alone while defining specific morphological targets for future domain adaptation.

\end{abstract}

\section{Introduction}
\label{sec:intro}

Prostate cancer is among the most diagnosed cancers in men, with 255,395 new cases reported in the United States in 2022 and 33,881 deaths in 2023 \cite{cdc2025prostate}. The diagnostic criterion standard requires tissue core biopsy followed by hematoxylin and eosin (H\&E) dye staining and microscopic histopathological examination by trained pathologists \cite{ravery2008twenty}. H\&E staining reveals nuclear and cytoplasmic structures essential for tumor identification and Gleason tumor grading, with up to three million slides stained daily worldwide \cite{feldman2014tissue}. In routine pathology workflows, histopathologic review is typically performed on an H\&E-stained section, whereas immunohistochemistry and molecular assays are often performed on adjacent unstained serial sections, commonly separated by approximately 4~\textmu m\cite{ravery2008twenty}\cite{feldman2014tissue}\cite{mccann2015automated}. Consequently, structural morphology and molecular readouts are not perfectly linked at the same microscopic location, which is particularly relevant for RNA-based assays that are sensitive to pre-analytic processing \cite{goldstein2010identification}. 

\subsection{Related Work}
Deep learning-based virtual staining has emerged as a promising alternative to chemical staining. Conditional generative adversarial networks (cGANs) can learn pixel-level mappings between unstained and H\&E-stained tissue whole slide images (WSI), simulating the staining process computationally without consuming physical tissue \cite{goodfellow2014generative}\cite{isola2017image}. There has been rapid growth in this field, with GAN-based frameworks, particularly U-Net architectures, being the most widely adopted \cite{bai2023deep}\cite{latonen2024virtual}. Virtual staining has been demonstrated across multiple tissue types and stain modalities \cite{bai2023deep}\cite{latonen2024virtual}, with recent advances including vision transformer and diffusion model frameworks \cite {klockner2025virtual}. The authors first reported cGAN-based computational H\&E staining and destaining of prostate core biopsy images at IEEE ICMLA 2018 \cite{rana2018computational}. In a comprehensive follow up study published in JAMA Network Open \cite{rana2020computational}, achieving Pearson correlation coefficient (PCC) of 0.962, structural similarity index measure (SSIM) of 0.902, and peak signal-to-noise ratio (PSNR) of 22.82 dB for H\&E staining of prostate core biopsy WSI, and PCC of 0.963, SSIM of 0.900, and PSNR of 25.65 dB for H\&E destaining, with approximately 95\% pixel-level agreement in tumor annotations in five board-certified pathologists. This research was further extended to an automated end-to-end framework combining computational staining with CNN-based tumor classification and weakly-supervised localization, demonstrating that computationally stained images can drive automated diagnostic pipelines without human review \cite{bayat2021automated}. In clinical practice, deployment on WSIs acquired at different institutions may introduce domain shift due to differences in tissue preparation, slide handling, scanner hardware, acquisition settings, and staining protocols \cite{niazi2019digital}. Standard stain normalization alone may be insufficient to eliminate site-specific artifacts \cite{flex2025knowledge}. Most virtual staining models are trained and evaluated on spatially registered image pairs collected at a single site \cite{latonen2024virtual}, whereas pixel-level registration between unstained and stained scans is not typically available in routine clinical workflows. Consequently, performance reported on paired or spatially registered single-institution datasets may overestimate real-world generalizability. Although recent studies have begun to address these challenges through architectural modifications  \cite{flex2025knowledge}\cite{ma2026misalignment}, the extent to which previously trained virtual staining models can be applied to WSIs acquired at different hospitals without retraining remains unclear.


\subsection{Summary of Contributions}

This study evaluated the cross-site robustness of the computational H\&E staining and destaining cGAN models previously reported by us in JAMA Network Open study {\cite{rana2020computational} using an external dataset of 82 previously unseen prostate core biopsy WSI acquired at Stanford University (dataset\_S). The image comparisons and evaluation methodologies used in this study are summarized in Table~\ref{tab1}. Unlike the original Brigham and Women’s Hospital training set (dataset\_B), the WSIs from dataset\_S were spatially unregistered and exhibited visible cross-site distribution shift. Rather than retraining the models, the study investigated whether usable performance could be recovered using preprocessing-based domain adaptation of input WSIs alone. Specifically, histogram-based stain normalization was applied to H\&E WSI and custom channel-wise intensity calibration to unstained WSI from dataset\_S, before inferencing with virtual H\&E staining and destaining models trained on dataset\_B. Registration on input dataset\_S WSI was intentionally not performed to simulate an out-of-domain evaluation scenario and to assess performance under minimal intervention conditions. The contributions of this paper are:
1. Demonstrate cross-site inference of previously trained cGAN-based H\&E staining and destaining models (trained on dataset\_B) on unseen prostate biopsy images in dataset\_S without retraining, using only preprocessing-based adaptation.

2. Provides a quantitative evaluation across multiple image similarity metrics, including PCC, SSIM, PSNR, and mean squared error (MSE), under intentionally unregistered conditions, thereby establishing conservative lower-bound estimates of real-world performance of generative deep learning models. 

3. Shows that H\&E restaining from computationally destained outputs outperforms direct staining from ground-truth unstained inputs, suggesting that input harmonization and preprocessing quality may be more limiting than generative capacity in the cross-site setting studied here. 

4. Reports pathologist-guided qualitative analysis showing generally preserved benign glandular architecture alongside morphological degradation in malignant glands.

5. Outlines a H\&E destain-restain digital loop for WSI as an indirect validation strategy for settings in which paired ground-truth unstained tissue is unavailable.

\vspace{0.3cm}
\begin{figure*}[htbp]
\centering

\begin{minipage}{\textwidth}
\centering

\begin{tikzpicture}[
    node distance=0.6cm,
    box/.style={
        draw,
        rounded corners,
        minimum width=1.5cm,
        minimum height=1.0cm,
        align=center,
        font=\footnotesize
    },
    arrow/.style={->, thick, shorten >=2pt, shorten <=2pt}
]

\node[box] (b1) {Dataset\_S\\Raw WSI};
\node[box, right=of b1] (b2) {Core\\Extraction};
\node[box, right=of b2] (b3) {Downsample};
\node[box, right=of b3] (b4) {Stain Norm\\+ Hist. Align};
\node[box, right=of b4] (b5) {Patch\\Extraction};
\node[box, right=of b5] (b6) {Destaining/\\Staining GAN};
\node[box, right=of b6] (b7) {Core\\Reconstruction};
\node[box, right=of b7] (b8) {Evaluation};

\draw[arrow] (b1) -- (b2);
\draw[arrow] (b2) -- (b3);
\draw[arrow] (b3) -- (b4);
\draw[arrow] (b4) -- (b5);
\draw[arrow] (b5) -- (b6);
\draw[arrow] (b6) -- (b7);
\draw[arrow] (b7) -- (b8);

\end{tikzpicture}

\vspace{3pt}
{\footnotesize (a) End to end computational destaining and restaining pipeline}

\end{minipage}

\vspace{10pt}

\begin{minipage}{\textwidth}
\centering

\begin{tikzpicture}[
    node distance=0.7cm,
    img/.style={
        draw,
        minimum width=1.6cm,
        minimum height=1.3cm,
        align=center,
        font=\scriptsize
    },
    model/.style={
        draw,
        minimum width=1.4cm,
        minimum height=0.9cm,
        align=center,
        font=\scriptsize
    },
    arrow/.style={->, thick, shorten >=2pt, shorten <=2pt}
]

\node[inner sep=0pt] (stained1)
{\includegraphics[width=1.6cm]{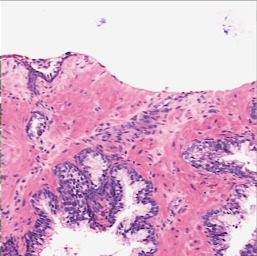}};
\node[font=\scriptsize, below=2pt of stained1] {GH\&E Input};
\node[model, right=of stained1] (dgan) {Destaining GAN};
\node[inner sep=0pt, right=of dgan] (destained1) {\includegraphics[width=1.6cm]{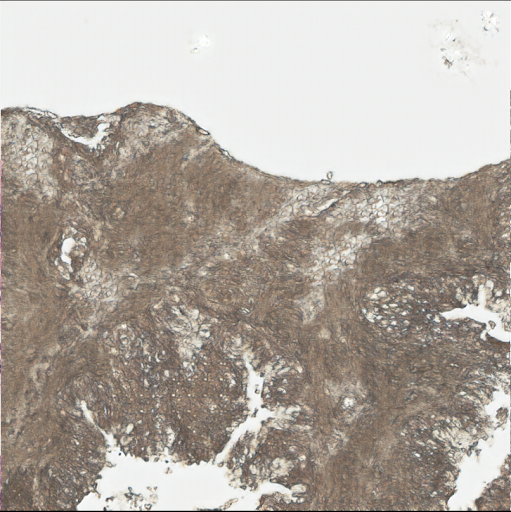}};
\node[font=\scriptsize, below=2pt of destained1] {VDS Output};

\node[inner sep=0pt, right=1.5cm of destained1] (unstained1) {\includegraphics[width=1.6cm]{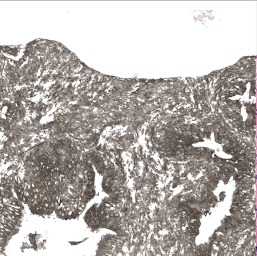}};
\node[font=\scriptsize, below=2pt of unstained1] {GUS Input};
\node[model, right=of unstained1] (sgan) {Staining GAN};
\node[inner sep=0pt, right=of sgan] (stained2) {\includegraphics[width=1.6cm]{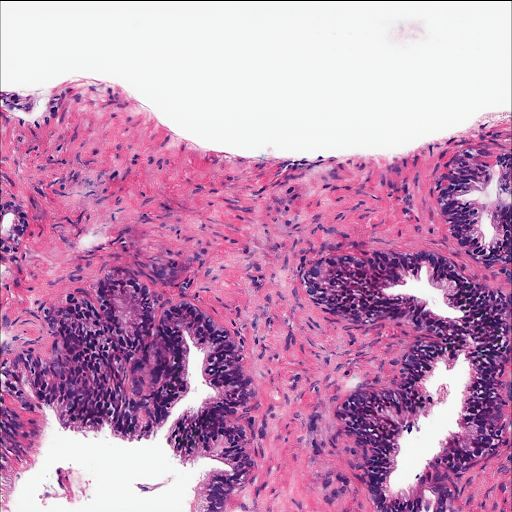}};
\node[font=\scriptsize, below=2pt of stained2] {VH\&E Output};

\draw[arrow] (stained1) -- (dgan);
\draw[arrow] (dgan) -- (destained1);
\draw[arrow] (unstained1) -- (sgan);
\draw[arrow] (sgan) -- (stained2);

\node[inner sep=0pt, below=1.4cm of dgan] (stainL) {\includegraphics[width=1.6cm]{images/stained_input.png}};
\node[font=\scriptsize, below=2pt of stainL] {GH\&E Input};
\node[model, right=of stainL] (dganL) {Destaining GAN};
\node[inner sep=0pt, right=of dganL] (destainL) {\includegraphics[width=1.6cm]{images/destained1.png}};
\node[font=\scriptsize, below=2pt of destainL] {VDS Output/Input};
\node[model, right=of destainL] (sganL) {Staining GAN};
\node[inner sep=0pt, right=of sganL] (restainL) {\includegraphics[width=1.6cm]{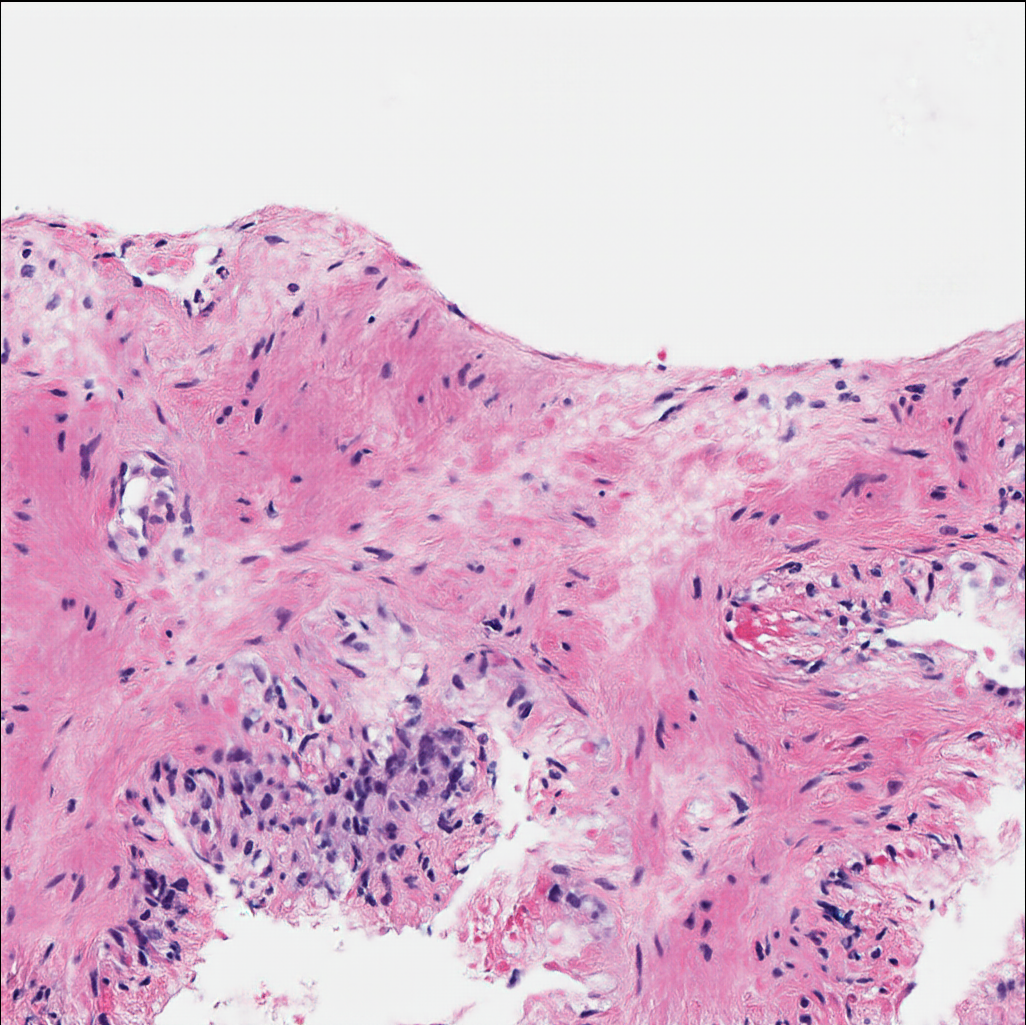}};
\node[font=\scriptsize, below=2pt of restainL] {VH\&ER Output};

\draw[arrow] (stainL) -- (dganL);
\draw[arrow] (dganL) -- (destainL);
\draw[arrow] (destainL) -- (sganL);
\draw[arrow] (sganL) -- (restainL);

\end{tikzpicture}

\vspace{3pt}
{\footnotesize (b) Three inference pathways denoting the input and output images for each model and the destaining-staining loop}

\end{minipage}

\caption{Computational destaining and restaining workflow. (a) End-to-end processing pipeline from raw whole-slide images (WSIs) in dataset\_S through core extraction, downsampling, preprocessing, GAN inference, reconstruction, and evaluation. (b) Transformation pathways used to generate each image type. In the destaining pathway, ground-truth H\&E WSIs (GH\&E) are input to the destaining GAN to generate virtually destained outputs (VDS). In the direct staining pathway, ground-truth unstained WSIs (GUS) are input to the staining GAN to generate virtually H\&E-stained outputs (VH\&E). In the integrated destain-restain pathway, GH\&E is first passed through the destaining GAN to generate VDS, which is then input to the staining GAN to produce virtually H\&E-restained outputs (VH\&ER).}
\label{fig1}

\end{figure*}
\vspace{-0.4cm}
\begin{table}[htbp]
\centering
\caption{Validation Framework for Computational Staining and Destaining. D, destaining GAN; S, staining GAN; D–S, sequential destaining followed by staining. Evaluation included pixel-wise metrics (PCC, SSIM, PSNR, MSE), RGB intensity differences, and expert pathologist review.}
\label{tab1}
\fontsize{7.5pt}{9.5pt}\selectfont 
\setlength{\tabcolsep}{3pt}
\renewcommand{\arraystretch}{1.5}
\begin{tabularx}{\columnwidth}{>{\hsize=0.53\hsize}X >{\centering\arraybackslash\hsize=0.4\hsize}X >{\hsize=1.07\hsize}X}
\textbf{Comparison} & \textbf{Model} & \textbf{Evaluation Description} \\ \hline

GUS vs. VDS & D & Destaining quality (PCC, SSIM, PSNR, MSE, RGB intensity differences). \\ [0.5ex]
GH\&Evs. VH\&ER & D--S & Destaining-staining loop quality (PCC, SSIM, PSNR, MSE, RGB intensity differences, pathologist review). \\ [0.5ex]
GH\&E vs. VH\&E & S & Direct staining quality (PCC, SSIM, PSNR, MSE, RGB intensity differences). \\ [0.5ex]
VH\&Evs. VH\&ER & D--S vs. S & Destaining-stage effect (PCC, SSIM, PSNR, MSE, RGB intensity differences). \\ 

Raw GUS vs. VDS & D & Model output vs. original tissue (Intensity RGB). \\ [0.5ex]
Raw GH\&E vs. VH\&ER & D--S & Model output vs. original tissue (intensity RGB). \\

GH\&E vs. VDS & D & Intensity change from destaining (Intensity RGB). \\ [0.5ex]
VH\&ER vs. VDS & D vs. D--S & Intensity change from restaining, staining loop consistency (intensity RGB). \\ 

 
  \hline
\end{tabularx}
\end{table}

\section{Data Description}
Prostate core biopsy WSIs were obtained from three patients at Stanford University under IRB protocol no. 53704, and deidentfied WSIs were transferred to the University of California, Irvine for analysis. Multiple biopsy cores from different patients were mounted on the same glass slide and separated by lane markers. This dataset\_S comprised 56 cases across the three patients, with each case containing one or two biopsy cores, yielding a total of 82 biopsy cores used for the quantitative evaluation. Each core was initially digitized as a ground-truth unstained (GUS) deparaffinized tissue WSI using a Leica Aperio AT2 scanner (Leica Biosystems, Vista, CA, USA). Following the initial scan, the same physical sections underwent chemical H\&E staining using a Leica HistoCore SPECTRA ST automated slide stainer(Leica Biosystems Nussloch GmbH, Nussloch, Germany). The slides were then rescanned to obtain corresponding ground-truth H\&E (GH\&E) images. All WSIs were obtained at an initial 40x magnification (0.16 µm/pixel) and subsequently exported at a 20x magnification equivalent. This workflow ensured that each image pair represents the identical physical tissue section captured before and after chemical staining. The pretrained cGAN models were originally trained on 102 registered dataset\_B WSIs, as reported in the prior JAMA study \cite{rana2020computational}. For the present study, baseline re-inference was performed on the 13 validation WSIs from dataset\_B without model retraining. The 112 unstained WSIs from 38 patients were scanned with the Aperio ScanScope XT system (Leica Biosystems, Buffalo Grove, IL, USA) at 20x magnification. Subsequently, the same slides were stained with H\&E dye on the Agilent Dako Autostainer (Agilent, Santa Clara, CA), and these stained slides were re-scanned on the Aperio ScanScope XT at 20x magnification at Harvard Medical School Tissue Microarray \& Imaging Core.

\section{Data Preprocessing}

\subsection{Core Extraction}

\newcommand{\inlineTilde}{\raisebox{-0.5ex}{\texttt{\char`~}}}

Individual biopsy cores were extracted from multi-core WSIs using QuPath \cite{bankhead2017qupath}. For each core, a GeoJSON region of interest (ROI) was manually annotated on the corresponding. svs image, and the enclosed tissue region was extracted using an automated script adapted from prior work \cite{rana2018computational} \cite{rana2020computational} . Separate GeoJSON annotations were generated for the unstained and H\&E-stained WSI scans of each core, since transferring annotations across scans could produce asymmetric boundaries at the pixel level due to differences in scanner positioning between scan sessions. The native resolution of the WSIs in the model inferencing dataset\_S was 0.1377 microns per pixel (MPP), with uncompressed file sizes exceeding 142 GB per slide. In contrast, the training data WSI in dataset\_B were acquired at 0.5 MPP. To match the resolution used during model training, all dataset\_S cores were downsampled to 0.5 MPP prior to subsequent processing. For very large cores, a strip-based extraction procedure was implemented in which the image was processed in horizontal strips of 10,000 pixels in width; this approach produced output equivalent to whole-core extraction while reducing memory requirements. The overall preprocessing workflow from raw dataset\_S WSIs through core extraction and downsampling is shown in Fig.~\ref{fig1}(a).

\subsection{Preprocessing: Stain Normalization Intensity Calibration}\label{sec_model_pred_eval}

Direct inference of dataset\_S WSIs using models trained on dataset\_B produced visibly degraded outputs, as illustrated in Fig.~\ref{fig2}. In particular, Fig.~\ref{fig2}(A-I) and Fig.~\ref{fig2}(B-I) show representative raw H\&E-stained and unstained input patches from dataset\_S, whereas Fig.~\ref{fig2}(A-III) and Fig.~\ref{fig2}(B-III) show the corresponding dataset\_B reference patches used for normalization and alignment. The effect of insufficient preprocessing is shown in Fig.~\ref{fig2}(C-I) and Fig.~\ref{fig2}(C-II), where a pre-normalization input patch yields a blurred model output. After preprocessing of the same input, Fig.~\ref{fig2}(C-III) shows a visibly improved output. To improve compatibility with the training distribution without retraining the models, two separate preprocessing strategies were developed for GH\&E and GUS images in dataset\_S. For GH\&E cores, stain normalization was first performed using the Macenko method implemented in TiaToolbox \cite {pocock2022tiatoolbox}. 

\begin{figure}[htbp]
\centering

\setlength{\tabcolsep}{1pt} 
\renewcommand{\arraystretch}{1} 

\begin{tabular}{c c c c}
 & \scriptsize I & \scriptsize II & \scriptsize III \\

\scriptsize (A) &
\includegraphics[height=2.3cm]{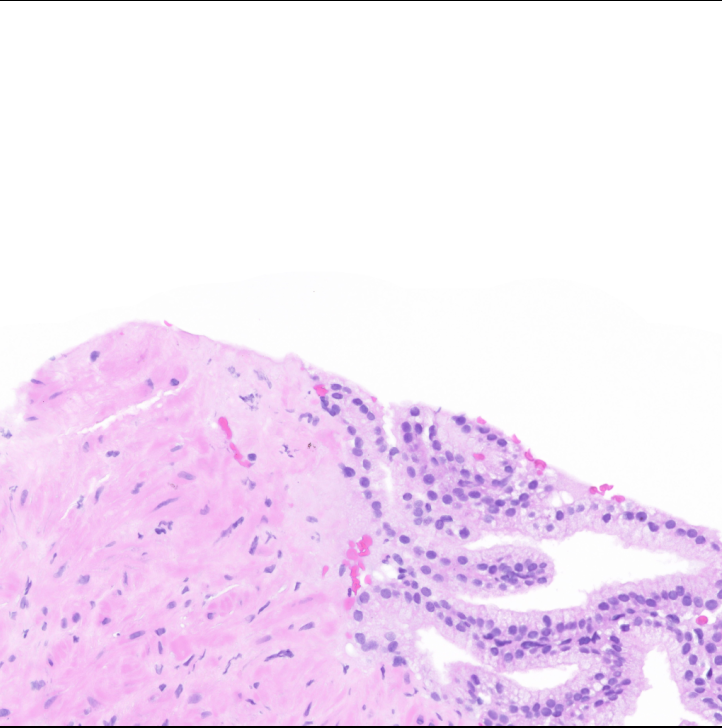} &
\includegraphics[height=2.3cm]{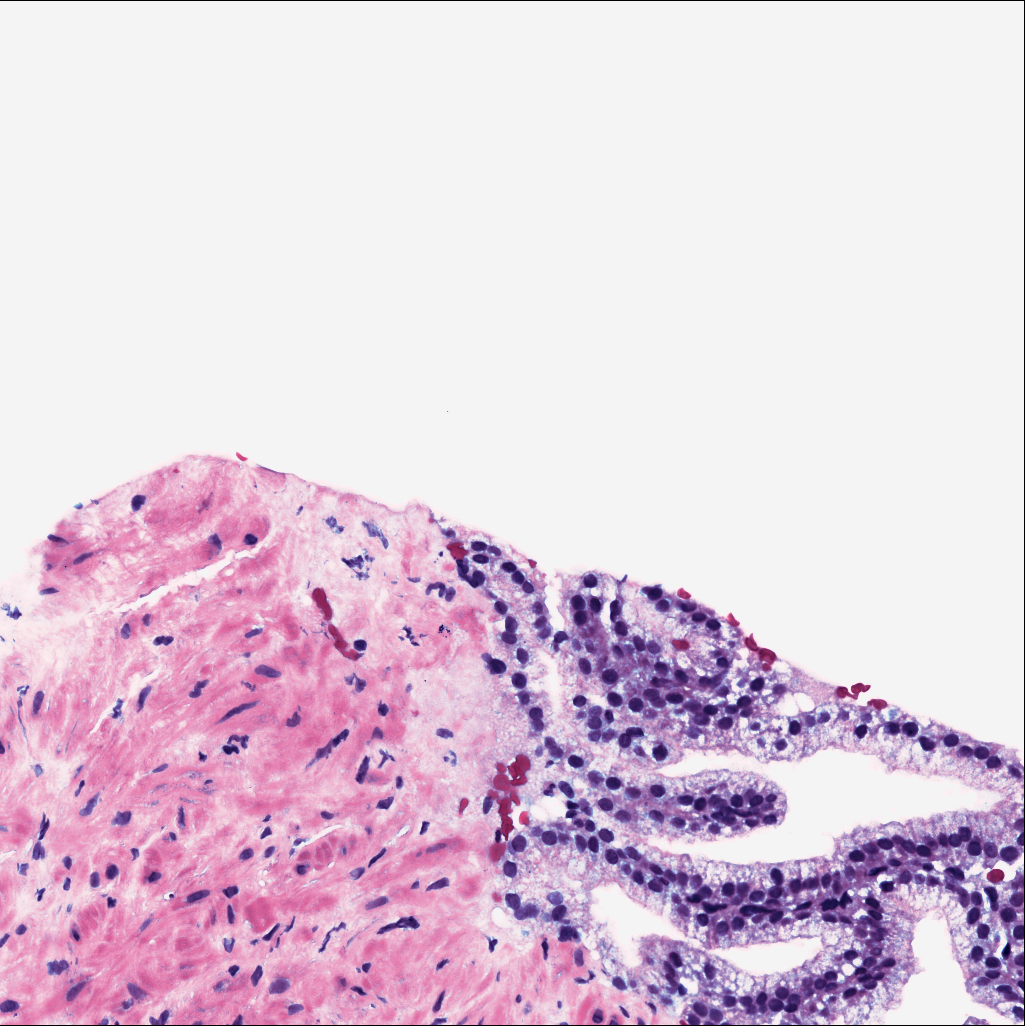} &
\includegraphics[height=2.3cm]{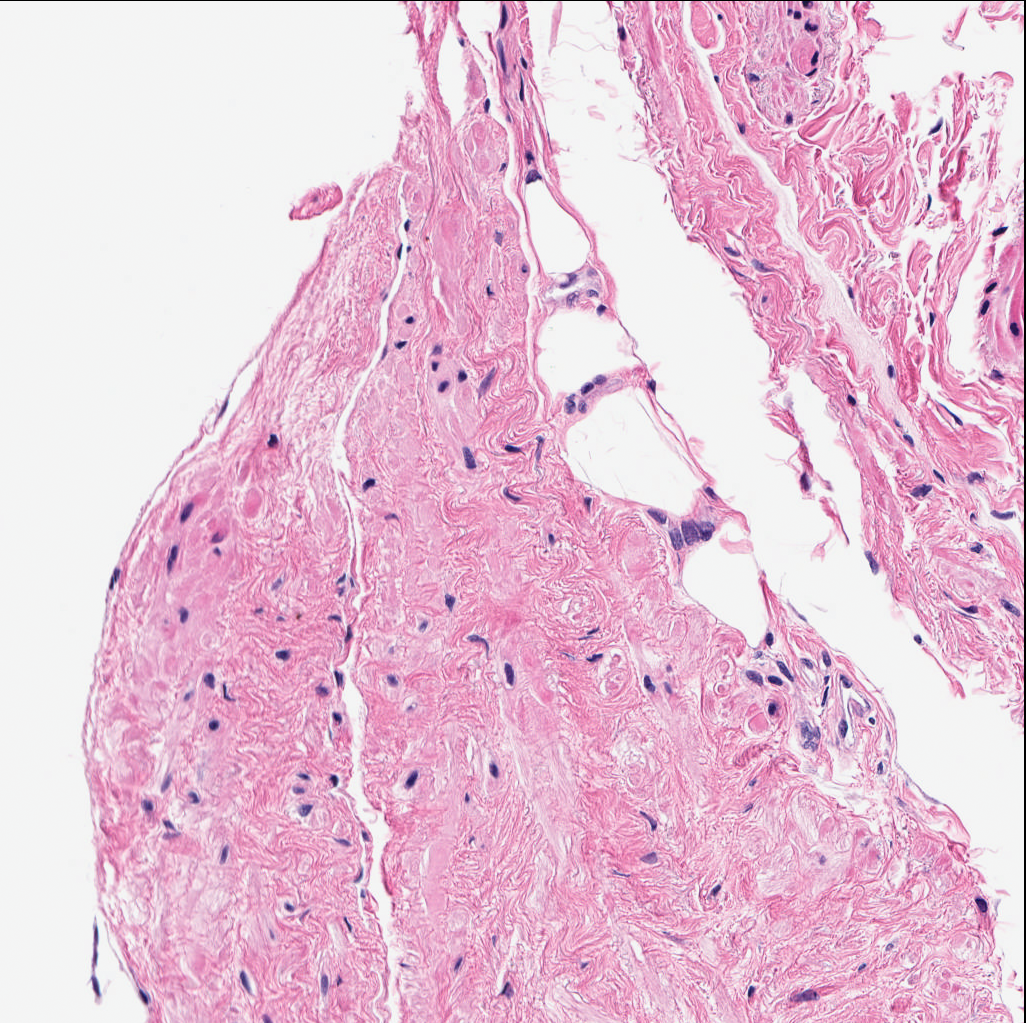} \\

\scriptsize (B) &
\includegraphics[height=2.3cm]{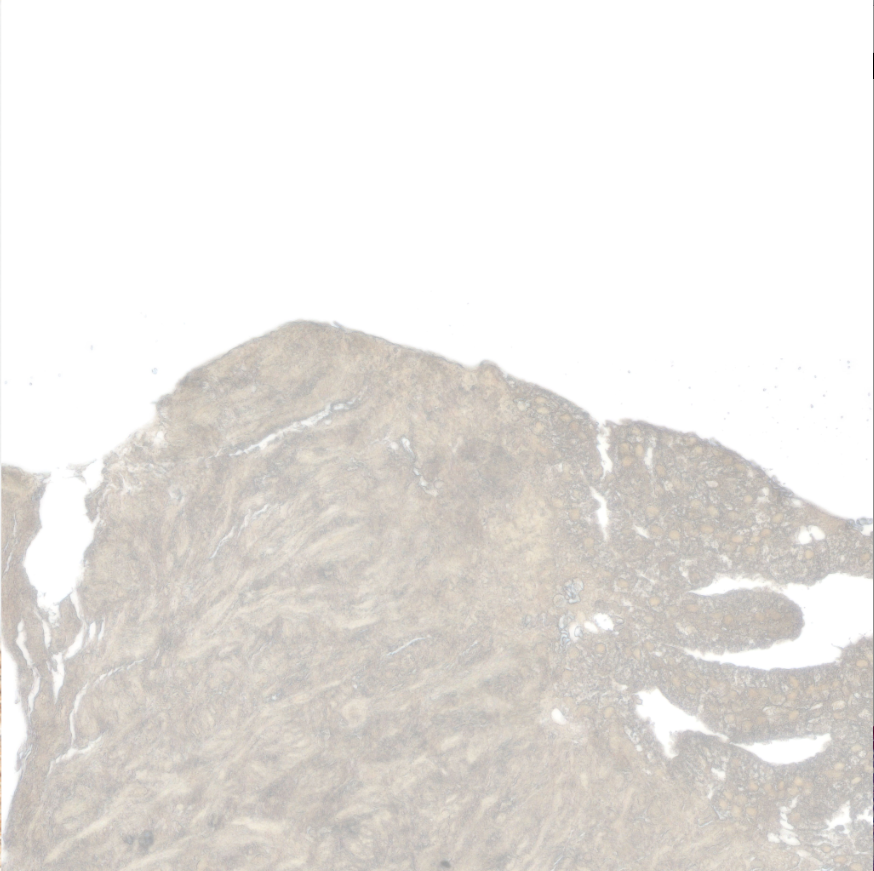} &
\includegraphics[height=2.3cm]{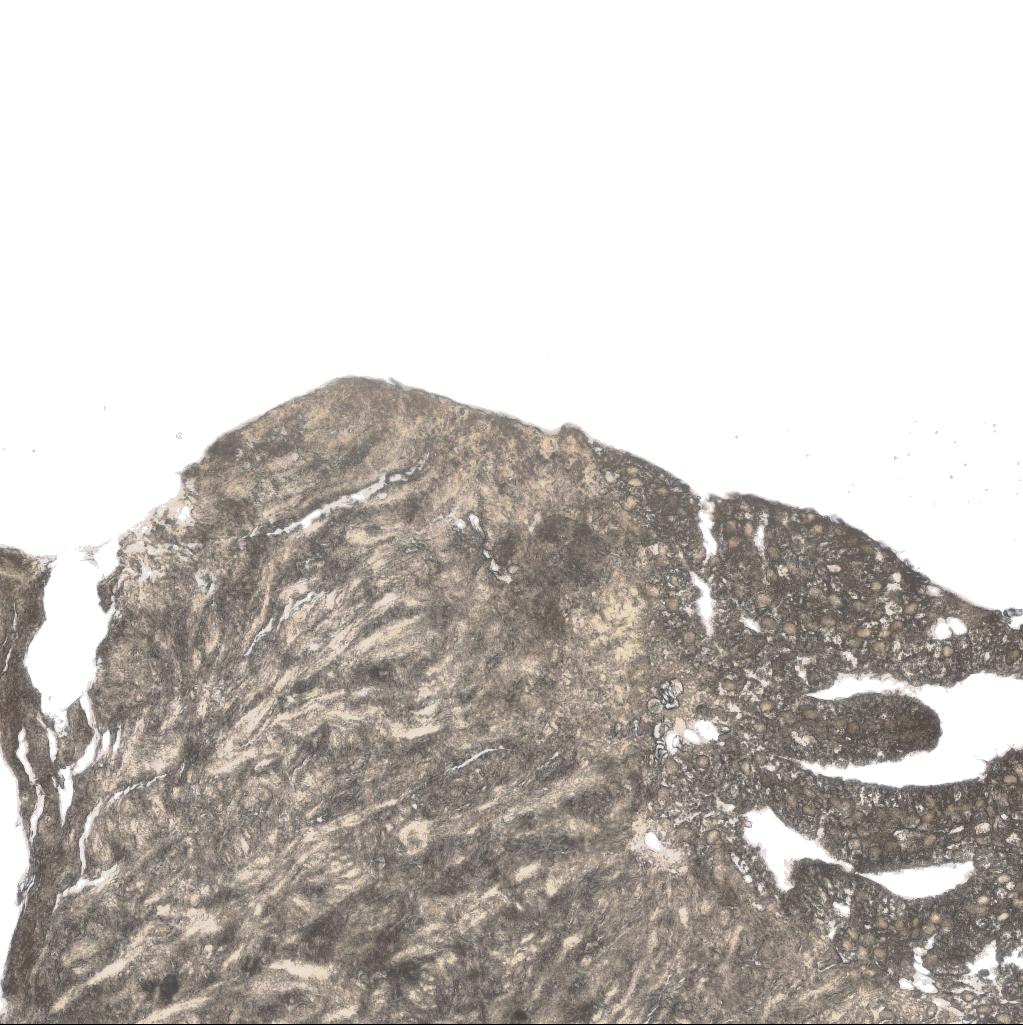} &
\includegraphics[height=2.3cm]{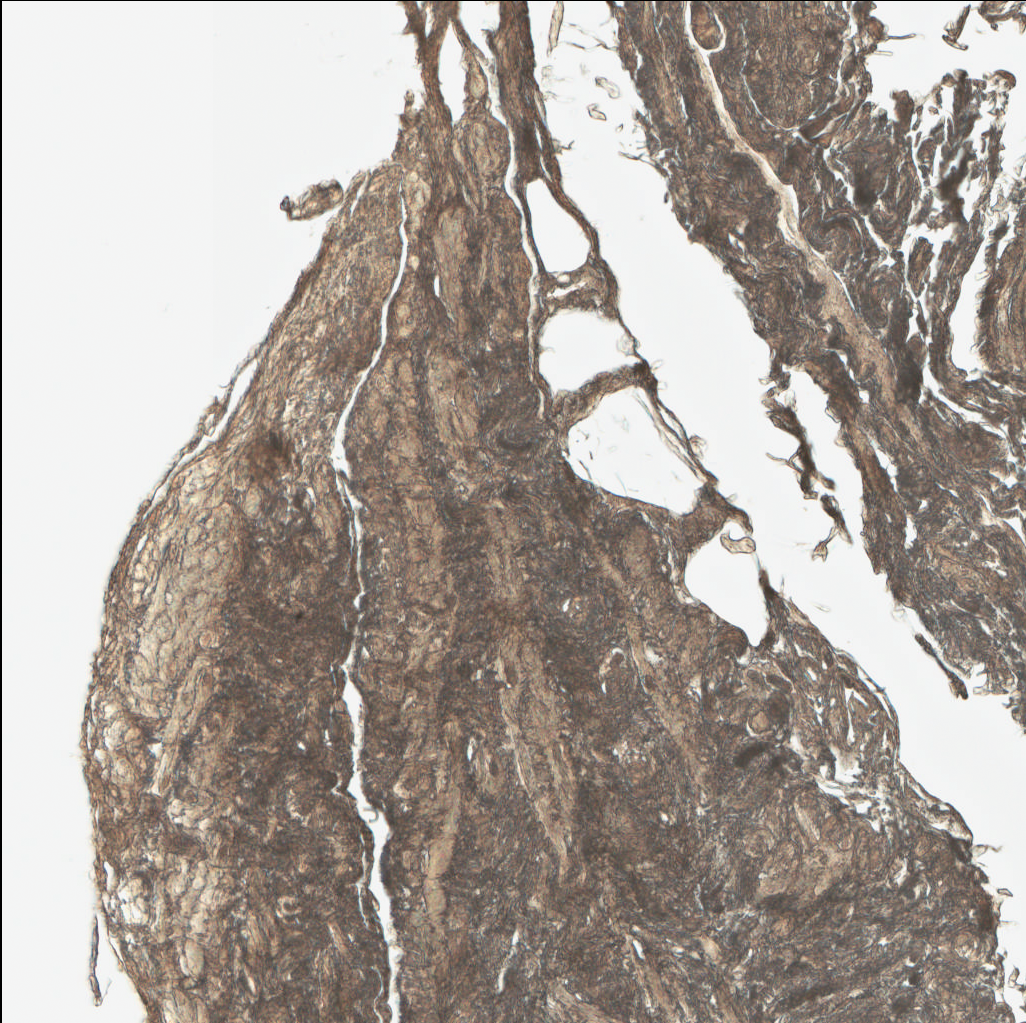} \\

\scriptsize (C) &
\includegraphics[height=2.3cm]{images/raw_stanford_input.png} &
\includegraphics[height=2.3cm]{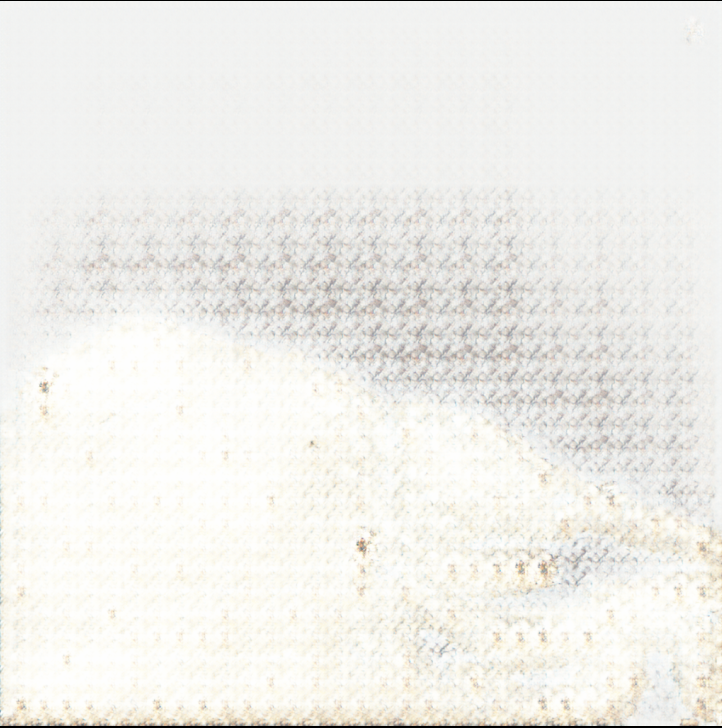} &
\includegraphics[height=2.3cm]{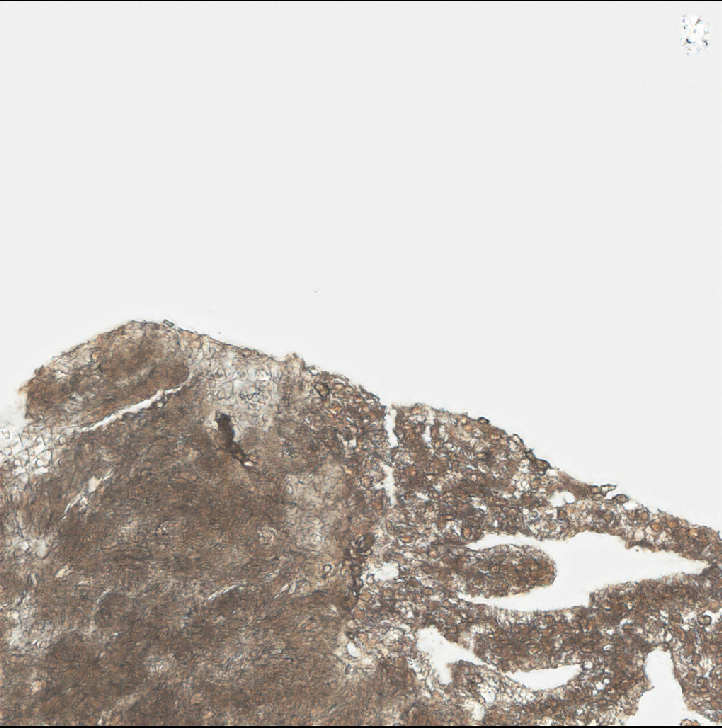} \\

\end{tabular}

\caption{Effect of preprocessing on inference performance after patch extraction. Individual patches are indexed by row (A–C) and column (I–III). Row A illustrates H\&E stain normalization and histogram alignment on raw H\&E-stained patches from dataset\_S. Row B illustrates normalization of raw unstained patches from dataset\_S. In Row C, C-I is a pre-normalization input patch, C-II is the corresponding blurred model output without sufficient preprocessing, and C-III is the improved output obtained after preprocessing the same input. A-III and B-III are reference patches from dataset\_B used for normalization and alignment.}
\label{fig2}

\end{figure}

This was followed by histogram alignment using cumulative distribution function matching to a representative GH\&E reference image from dataset\_B. A tissue mask was applied during histogram alignment so that only tissue-containing regions contributed to the transformation, thereby preventing background whitespace from distorting the color mapping. This two-step procedure addressed both stain-vector variation and global intensity-distribution mismatch between dataset\_S and dataset\_B WSIs (Fig.~\ref{fig2}). For non-stained GUS cores, standard stain normalization approaches were not applicable because they are designed for chemically stained tissue images. Therefore, a custom intensity calibration procedure was developed based on histogram matching with modified channel-wise RGB weighting. The channel weights were iteratively tuned to match the visual appearance of GUS reference images from dataset\_B. This procedure adjusted the color balance and saturation of dataset\_S unstained images to better approximate the input distribution of dataset\_B WSIs seen during training.

\vspace*{-0.2cm}
\subsection{Patch Extraction} \label{sec:results_and_disc3}
After preprocessing, each core image was divided into non-overlapping patches of size 1024 × 1024 pixels, matching the input size of the pretrained cGAN models. Patches were extracted on a regular grid, and zero-padding was applied at image boundaries when tissue did not completely occupy a full patch. To avoid processing regions containing minimal tissue, patches with predominantly background content were excluded using a tissue-content threshold.

\section{Methods} \label{sec:model_inference_reconstruction}

\subsection{Model Architecture} \label{sec:model_inference_reconstruction1}
Pretrained cGAN models for computational H\&E staining and destaining, previously developed on paired, spatially registered WSIs from dataset\_B, were used without retraining, fine-tuning, or architectural modification \cite{rana2018computational} \cite{rana2020computational} \cite{bayat2021automated}. Both models used the pix2pix framework \cite{isola2017image} with a U-Net generator (1024 × 1024 × 3 input/output) and PatchGAN discriminator, trained with adversarial, L1, and PCC losses. Full architecture and training details are described in \cite{rana2018computational} \cite{rana2020computational}. Inference was performed patchwise on an NVIDIA GeForce RTX 4080 GPU.

\subsection{Model Inference Workflow} \label{sec:model_inference_reconstruction2}
Three inference pathways were evaluated on dataset\_S (Fig.~\ref{fig1}(b)):

\begin{enumerate}

    \item \textbf{Destaining:} Preprocessed GH\&E patches were passed through the H\&E destaining model to generate VDS patches (GH\&E $\rightarrow$ destaining model $\rightarrow$ VDS).

    \item \textbf{Direct staining:} Preprocessed GUS patches were passed through the H\&E staining model to generate VH\&E patches (GUS $\rightarrow$ staining model $\rightarrow$ VH\&E).

    \item \textbf{Restaining:} VDS patches generated in the first pathway were subsequently passed through the staining model to produce virtually H\&E-restained (VH\&ER) patches. This third pathway was designed to evaluate a destain-restain digital loop (GH\&E $\rightarrow$ destaining model $\rightarrow$ VDS $\rightarrow$ staining model $\rightarrow$ VH\&ER).

\end{enumerate}

For all three pathways, reconstructed whole-core outputs were assembled from the generated patches using their original grid coordinates.

\subsection{Quantitative Evaluation} \label{sec:model_inference_reconstruction3}
1) \textbf{Pixel-level similarity metrics}: PCC, SSIM, PSNR, and MSE were used for each comparison. Before metric computation, image pairs were aligned using a rigid transformation estimated with the Enhanced Correlation Coefficient (ECC) algorithm implemented in OpenCV (cv2.findTransformECC). This procedure optimized translation and rotation to maximize correlation between WSI pairs. Because only rigid alignment was applied, the registration did not fully correct for tissue deformation or local distortions introduced during staining and rescanning. Accordingly, all reported pixel-level metrics should be interpreted as conservative lower-bound estimates of model performance rather than values obtained under fully registered conditions. Four image comparisons were evaluated: VDS vs. GUS: virtual destaining quality; VH\&ER vs. GH\&E: restaining quality; VH\&E vs. GH\&E: direct staining quality; VH\&ER vs. VH\&E: effect of the destain-restain pathway relative to direct staining (n = 81). \\
2) \textbf{Pixel intensity analysis}: Mean pixel intensity was computed for each WSI after applying a tissue mask to exclude background regions. Intensity differences were then calculated both overall, using mean RGB intensity, and separately for each color channel (R, G, and B) for each comparison pair. Tissue masking was necessary because inclusion of background whitespace systematically increased average intensity values and distorted inter-image comparisons. \\
3) \textbf{Residual dataset\_B–dataset\_S domain shift}: To quantify residual domain shift after preprocessing, intensity differences were computed between each of the 82 normalized GH\&E cores from dataset\_S and 112 GH\&E training images from dataset\_B. For each dataset\_s inference core, the mean and median intensity difference relative to the dataset\_B training distribution were calculated as summary measures of residual cross-site discrepancy.

\subsection{Clinical Assessment} \label{sec:model_inference_reconstruction4}
Computationally restained whole-core VH\&ER images were qualitatively evaluated by a board-certified anatomic pathologist. The pathologist examined paired GH\&E images and corresponding VH\&ER images for four representative cases spanning a range of tumor burden and Gleason patterns: Case 44, cores 12 and 6 (Gleason grade 3); Case 42, core 10 (Gleason 3+4 with cribriform and poorly formed patterns); and Case 32, core 4 (predominantly Gleason grade 3 with focal grade 4). For each case, the pathologist assessed preservation of benign glandular architecture, fidelity of malignant gland morphology, maintenance of Gleason grading patterns, spatial correspondence of glands and stroma, and overall diagnostic utility. 

\subsection{Statistical Analysis} \label{sec:model_inference_reconstruction5}
Statistical comparisons among image similarity metrics were performed using analysis of variance (ANOVA) followed by Fisher’s least significant difference (LSD) post hoc testing. A \textit{p} value less than 0.05 was statistically significant. 

\section{Results} \label{sec:results_and_disc}
We evaluated the cross-site robustness of the computational H\&E staining and destaining cGAN models previously reported in our JAMA Network Open study [12] using an external dataset of 82 previously unseen prostate core biopsy WSIs (dataset\_S). In contrast to the original training set (dataset\_B), dataset\_S consisted of spatially unregistered WSIs with visible cross-site distribution shift. Table~\ref{tab2} summarizes pixel-level performance across the four principal image-comparison groups. Among the previously unseen dataset\_S WSI, virtual destaining comparison against normalized ground-truth unstained tissue (VDS vs. GUS) achieved the highest PCC, 0.854 ± 0.04, indicating the strongest global correspondence (Table~\ref{tab2}). This comparison also showed the lowest SSIM, 0.699 ± 0.11, indicating a potential dissociation between global pixel correspondence and local structural similarity under cross-site deployment. The corresponding PSNR (18.41 ± 1.55 dB) and MSE (0.015 ± 0.006) indicate moderate reconstruction fidelity (Table~\ref{tab2}).

\begin{table}[!t]
\centering
\caption{\textsc{Mean quantitative results for the 82 WSI core-level comparisons in the external validation cohort (Dataset\_S). The bottom two rows report mean baseline performance from re-inference on the 13 validation WSIs from Dataset\_B, derived from the validation set used in the previously reported JAMA study \cite{rana2020computational}.}}

\label{tab2}
\fontsize{7pt}{8.5pt}\selectfont
\setlength{\tabcolsep}{2pt}
\renewcommand{\arraystretch}{1.35}

\begin{tabularx}{\columnwidth}{
>{\raggedright\arraybackslash}X
>{\centering\arraybackslash}p{1.15cm}
>{\centering\arraybackslash}p{1.15cm}
>{\centering\arraybackslash}p{1.2cm}
>{\centering\arraybackslash}p{1.05cm}
}
\hline
\textbf{Comparison} & \textbf{PCC} & \textbf{SSIM} & \textbf{PSNR} & \textbf{MSE} \\ \hline

GUS vs. VDS
& $0.854\,\pm\,0.04$
& $0.70\,\pm\,0.11$
& $18.4\,\pm\,1.55$
& $0.015\,\pm\,0.00$ \\

GH\&E vs. VH\&ER
& $0.798\,\pm\,0.03$
& $0.756\,\pm\,0.09$
& $20.08\,\pm\,1.82$
& $0.010\,\pm\,0.01$ \\

GH\&E vs. VH\&E
& $0.715\,\pm\,0.05$
& $0.718\,\pm\,0.11$
& $18.5\,\pm\,1.96$
& $0.015\,\pm\,0.01$ \\

VH\&E vs. VH\&ER
& $0.733\,\pm\,0.05$
& $0.716\,\pm\,0.11$
& $19.21\,\pm\,2.24$
& $0.013\,\pm\,0.01$ \\

Dataset\_B* GH\&E vs. Dataset\_B VH\&ER
& $0.87\,\pm\,0.03$
& $0.86\,\pm\,0.07$
& $23.76\,\pm\,2.55$
& $0.005\,\pm\,0.003$ \\

Dataset\_B* VDS vs. Dataset\_B GUS
& $0.95\,\pm\,0.03$
& $0.86\,\pm\,0.07$
& $25.34\,\pm\,2.62$
& $0.003\,\pm\,0.002$ \\ \\
\hline
\end{tabularx}
\vspace{0.2em}
{\footnotesize \textit{*Dataset\_B values were obtained by re-inference on 13 validation WSIs and may differ slightly from the previously reported values in \cite{rana2020computational}.}}

\end{table}

In contrast, restaining through the destaining-staining digital loop (VH\&ER vs. GH\&E) outperformed direct computational staining from ground-truth unstained tissue (VH\&E vs. GH\&E) across all four quantitative metrics (Table~\ref{tab2}). The restaining loop achieved higher PCC (0.798 ± 0.04 vs. 0.715 ± 0.06), higher SSIM (0.756 ± 0.09 vs. 0.718 ± 0.11), higher PSNR (20.08 ± 1.82 dB vs. 18.51 ± 1.96 dB), and lower MSE (0.011 ± 0.004 vs. 0.015 ± 0.006) (Table~\ref{tab2}). Thus, the two-stage destaining-restaining pathway produced H\&E outputs that were consistently closer to ground-truth H\&E than direct staining from normalized unstained input, despite requiring sequential inference through two models. This result suggests that the model-generated destained intermediate WSI was better aligned with the staining model’s learned input distribution than the heuristically normalized unstained reference images. 

The two virtual H\&E outputs (VH\&ER vs. VH\&E) showed intermediate similarity (PCC 0.733, SSIM 0.716), confirming that the restained output remained more faithful to ground-truth H\&E than the directly stained output. When each was compared to ground-truth H\&E, VH\&ER achieved significantly higher PCC than VH\&E (Fisher's LSD, $p < 0.001$), confirming that the destain-restain loop produces a subtly but significantly better-aligned output rather than a fundamentally different image.

Re-inference of the trained models on the 13 validation WSIs from dataset\_B produced higher performance than external deployment on dataset\_S, with PCC values of 0.95 for destaining and 0.87 for restaining, SSIM values of 0.86 for both, PSNR values of 25.34 dB and 23.76 dB, and MSE values of 0.003 and 0.005, respectively (Table~\ref{tab2}). Relative to these dataset\_B validation baselines, performance on dataset\_S was lower for both comparisons, with a larger PCC drop for destaining (0.096) than for restaining (0.072). These results indicate that the pretrained models retained meaningful performance on dataset\_S without retraining, while also showing residual loss of fidelity under external deployment

Table~\ref{tab3} reports masked tissue-level RGB intensity differences across image comparisons. Intensity differences are interpreted in the order listed in the table, such that each comparison is calculated as the first image minus the second. Positive values indicate channels that are brighter in the first WSI whereas negative values indicate channels that are brighter in the second. The destaining transformation from H\&E to virtual destained tissue WSI produced a near-zero overall intensity difference between GH\&E and VDS (−0.88 ± 8.48 units), but with high channel redistribution (Table~\ref{tab3}). Relative to VDS, GH\&E had higher red intensity (+19.04 units) and higher blue intensity (+14.24 units), whereas VDS had higher green intensity, reflected by the negative green-channel difference (−35.92 units) (Table~\ref{tab3}). This channel-redistribution pattern, observed when comparing GH\&E with VDS, is consistent with reduction of eosin-associated red/pink signal and hematoxylin-associated blue/purple signal, yielding the green-brown appearance of virtually unstained tissue. These results indicate that destaining operated primarily as a color transformation rather than as a large overall intensity transformation. The restaining transformation showed the complementary channel pattern. In the VH\&ER vs. VDS comparison, the restained VH\&ER increased red intensity by 25.55 units, decreased green intensity by 27.42 units relative to VDS, and increased blue intensity by 18.33 units (Table~\ref{tab3}). Together, these paired channel shifts indicate that the destaining and restaining stages applied systematic, approximately inverse color transformations while maintaining a relatively small overall intensity shift across the digital loop.

The largest intensity difference among the primary evaluation WSI pairs was observed for GUS vs. VDS, with an overall difference of −39.83 ± 9.13 units and similar shifts across red, green, and blue channels (R: −38.49, G: −40.50, B: −40.50) (Table~\ref{tab3}). This largely uniform channel gap suggests that the difference was driven primarily by global intensity normalization rather than by a channel-specific failure of the destaining model. The VDS output remained closer to the normalized H\&E input intensity space, whereas the normalized unstained reference WSI was shifted more aggressively toward a darker distribution. This also helps explain the PCC-SSIM dissociation observed for VDS vs. GUS, the model preserved strong global correspondence, while local luminance and contrast mismatches reduced SSIM. This interpretation is supported by the raw-intensity comparisons. For example, VDS WSIs from dataset\_S were 25.07 ± 11.21 intensity units darker than raw unstained WSIs from dataset\_S, with a relatively uniform channel gap (Table~\ref{tab3}). Similarly, VH\&ER WSIs were 33.52 ± 4.66 intensity units darker than raw H\&E WSIs from dataset\_S, with the largest deficit in the blue channel (47.55 units). Thus, both model-generated outputs occupied a lower-intensity color space closer to the dataset\_B training distribution than to the raw dataset\_S inference distribution.

\begin{table}[!t]
\centering
\caption{\textsc{Mean pixel intensity differences for the 82 WSI core-level comparisons in the external validation cohort (Dataset\_S). For each row, intensity differences are computed as the first image type minus the second image type (e.g., GUS vs. VDS corresponds to GUS − VDS). The bottom two rows report mean baseline intensity differences from re-inference on the 13 validation WSIs from Dataset\_B, derived from the validation set used in the previously reported JAMA study \cite{rana2020computational}.}}
\label{tab3}
\fontsize{7pt}{8.5pt}\selectfont
\setlength{\tabcolsep}{2pt}
\renewcommand{\arraystretch}{1.35}

\begin{tabularx}{\columnwidth}{
>{\raggedright\arraybackslash}X
>{\centering\arraybackslash}p{1.15cm}
>{\centering\arraybackslash}p{1.15cm}
>{\centering\arraybackslash}p{1.15cm}
>{\centering\arraybackslash}p{1.15cm}
}
\hline
\textbf{Comparison} & \textbf{Overall} & \textbf{R} & \textbf{G} & \textbf{B} \\ \hline

GUS vs. VDS
& $-39.8\,\pm\,9.13$
& $-38.5\,\pm\,8.16$
& $-40.5\,\pm\,9.34$
& $-40.5\,\pm\,9.93$ \\

GH\&E vs. VH\&ER
& $-6.37\,\pm\,2.82$
& $-6.50\,\pm\,2.47$
& $-8.50\,\pm\,4.32$
& $-4.09\,\pm\,2.75$ \\

GH\&E vs. VH\&E
& $-1.08\,\pm\,8.17$
& $0.33\,\pm\,7.13$
& $-2.28\,\pm\,10.86$
& $-1.31\,\pm\,6.82$ \\

VH\&E vs. VH\&ER
& $5.31\,\pm\,9.27$
& $6.83\,\pm\,7.42$
& $6.27\,\pm\,12.62$
& $2.81\,\pm\,8.06$ \\

Raw GUS vs. VDS
& $25.07\,\pm\,11.21$
& $22.50\,\pm\,10.39$
& $25.38\,\pm\,11.21$
& $27.35\,\pm\,12.10$ \\

Raw GH\&E vs. VH\&ER
& $33.52\,\pm\,4.66$
& $22.50\,\pm\,4.24$
& $30.47\,\pm\,7.04$
& $47.55\,\pm\,3.45$ \\

GH\&E vs. VDS
& $-0.88\,\pm\,8.48$
& $19.04\,\pm\,7.56$
& $-35.92\,\pm\,8.96$
& $14.24\,\pm\,9.05$ \\

VH\&ER vs. VDS
& $5.49\,\pm\,7.90$
& $25.55\,\pm\,8.41$
& $-27.42\,\pm\,7.66$
& $18.33\,\pm\,7.97$ \\

Dataset\_B VH\&ER vs. Dataset\_B GH\&E
& $-4.60\,\pm\,4.20$
& $-4.20\,\pm\,3.97$
& $-5.30\,\pm\,5.18$
& $-4.31\,\pm\,3.82$ \\

Dataset\_B VDS vs. Dataset\_B GUS
& $-0.2\,\pm\,0.10$
& $-0.22\,\pm\,0.10$
& $-0.20\,\pm\,0.10$
& $-0.20\,\pm\,0.10$ \\
\hline
\end{tabularx}
\end{table}
By contrast, the comparison between ground-truth H\&E and directly stained virtual H\&E WSIs in dataset\_S (GH\&E vs. VH\&E) showed the smallest overall intensity gap among the three primary ground-truth comparisons, with a mean difference of only -1.08 ± 8.17 units (Table~\ref{tab3}), which was not statistically significant (ANOVA, F = 1.36, p = 0.25). This indicates that the direct staining model reproduced the global H\&E intensity distribution relatively well, even though its pixel-level structural similarity to ground-truth H\&E WSIs was lower than that of the restaining loop. Accordingly, the main limitation of direct computational staining in this cross-site setting appears to be structural fidelity rather than global intensity mismatch.
\begin{figure}[htbp]
\centering

\setlength{\tabcolsep}{1pt} 
\renewcommand{\arraystretch}{1} 

\begin{tabular}{c c c c}
 & \scriptsize I & \scriptsize II & \scriptsize III \\

\scriptsize (A) &
\includegraphics[height=2.3cm, align=c]{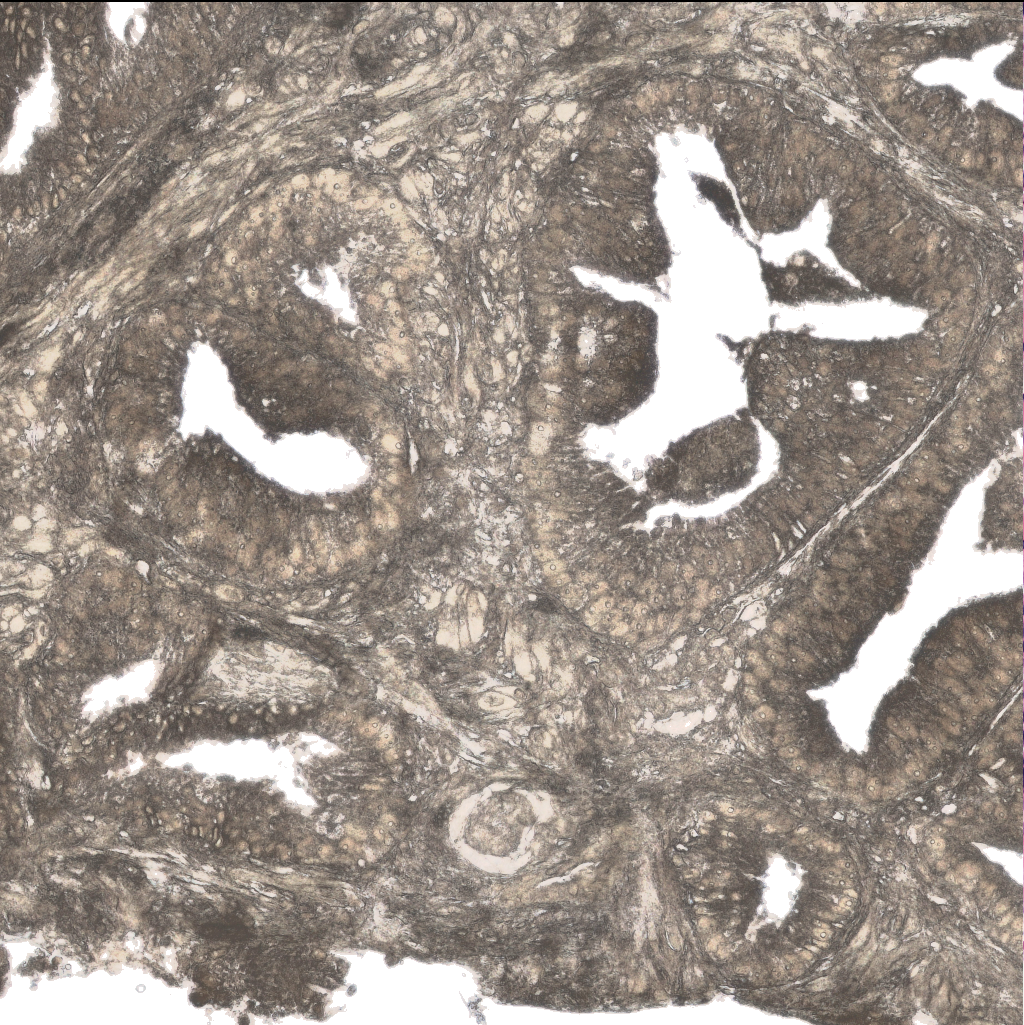} &
\includegraphics[height=2.3cm, align=c]{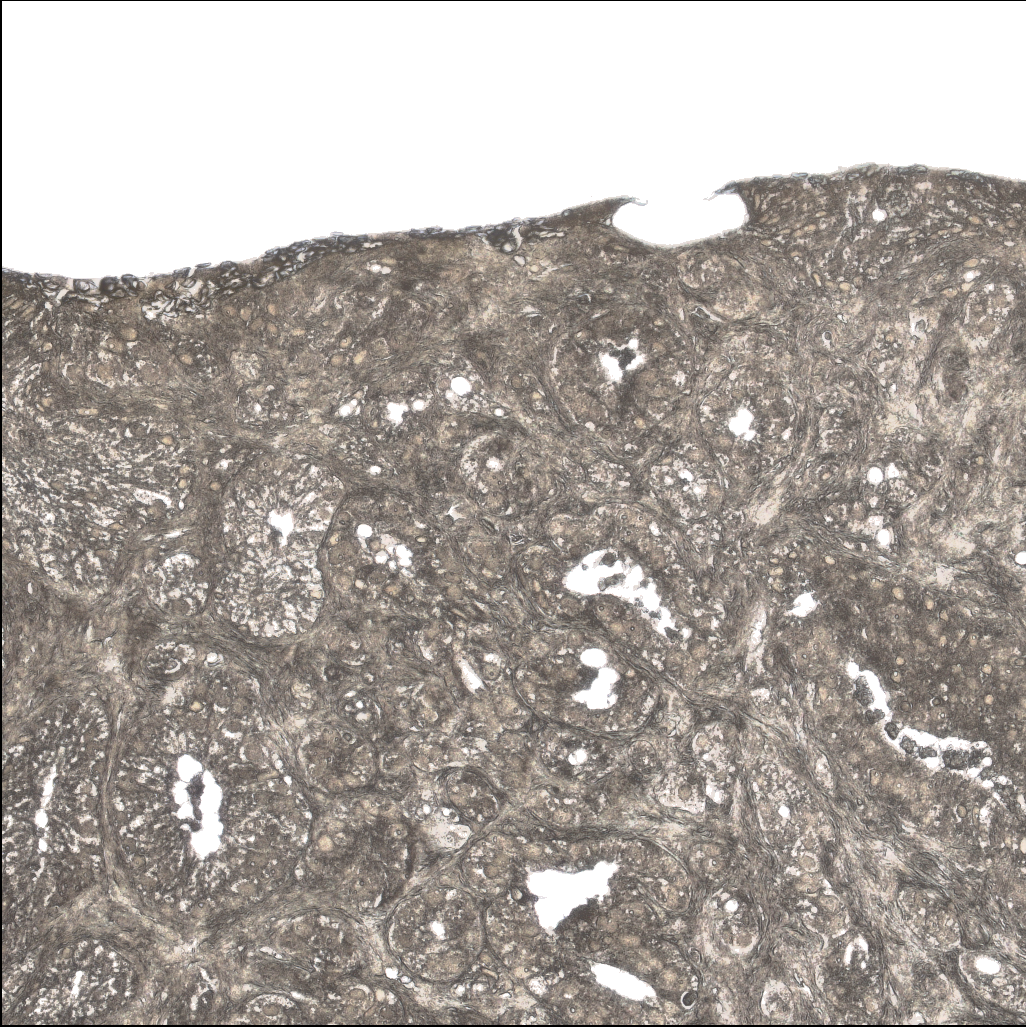} &
\includegraphics[height=2.3cm, align=c]{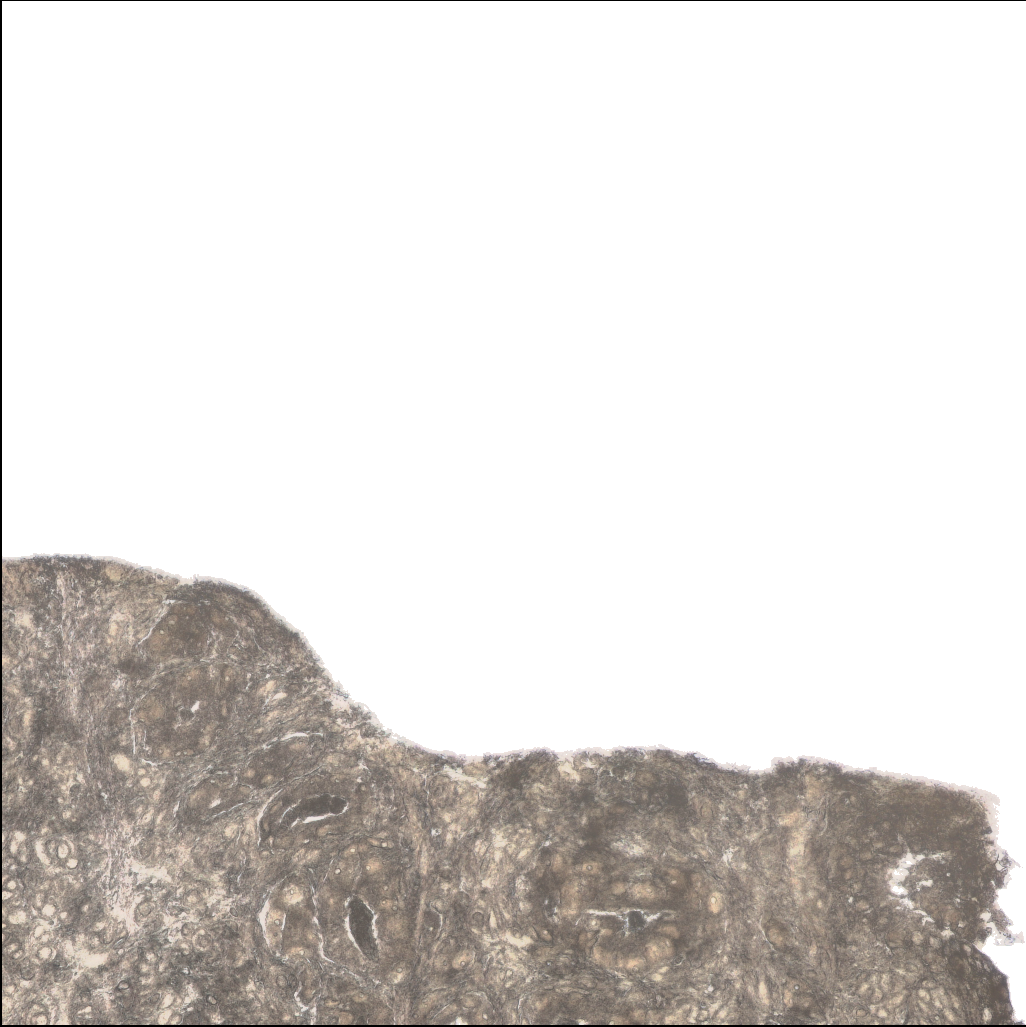} \\

\scriptsize (B) &
\includegraphics[height=2.3cm, align=c]{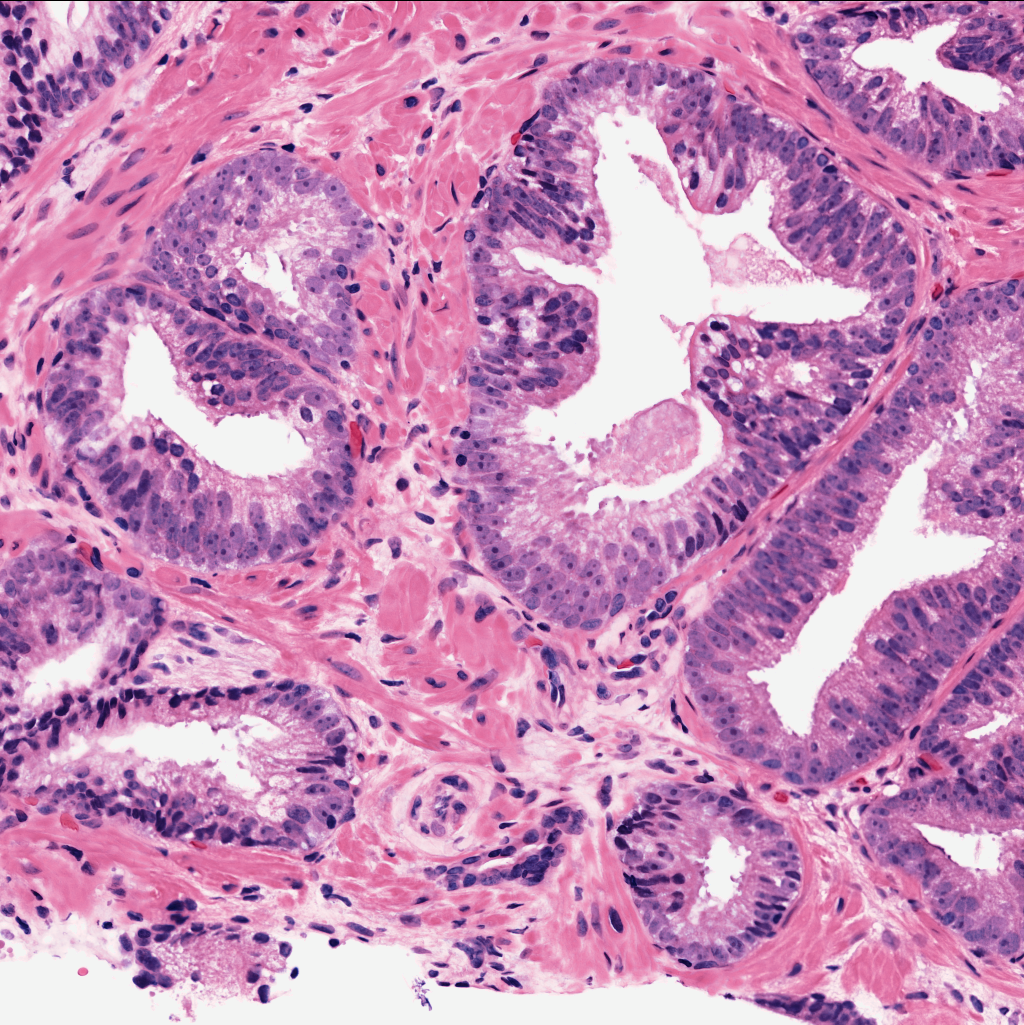} &
\includegraphics[height=2.3cm, align=c]{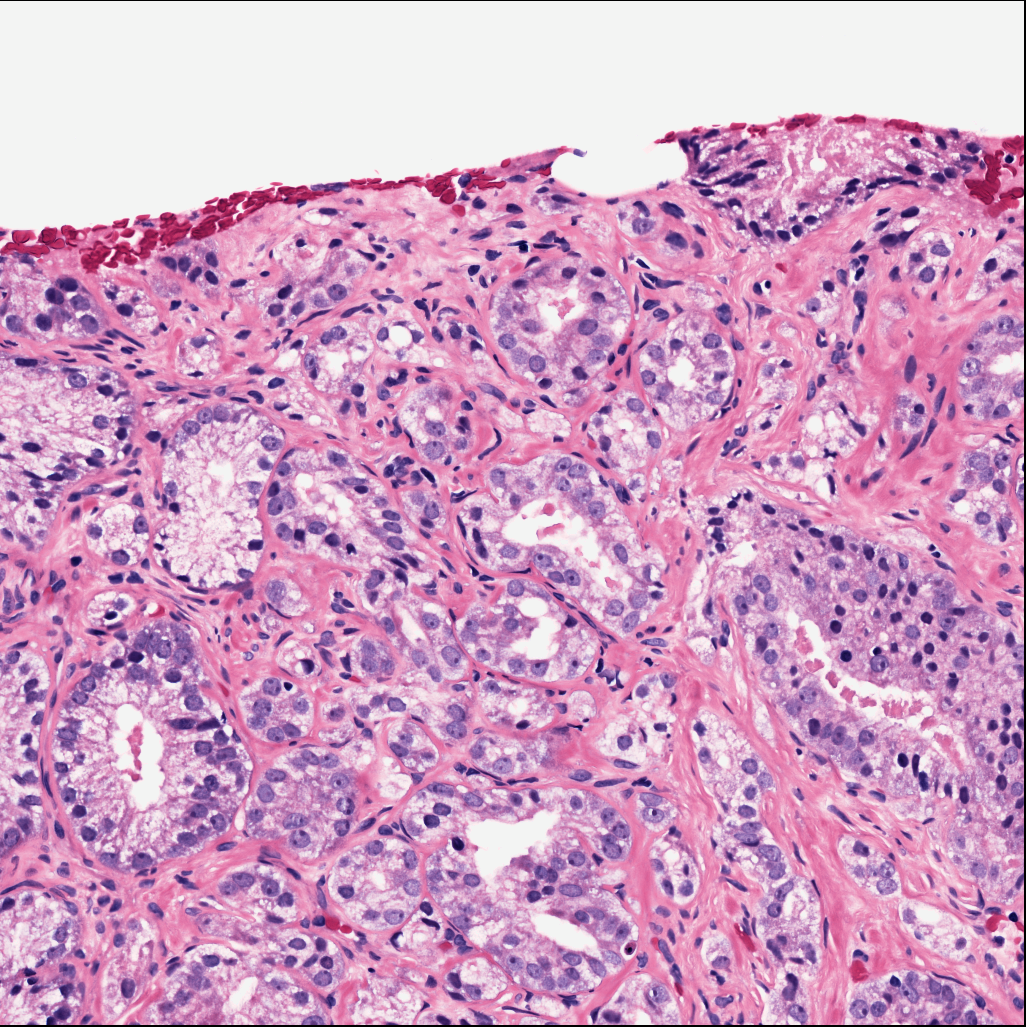} &
\includegraphics[height=2.3cm, align=c]{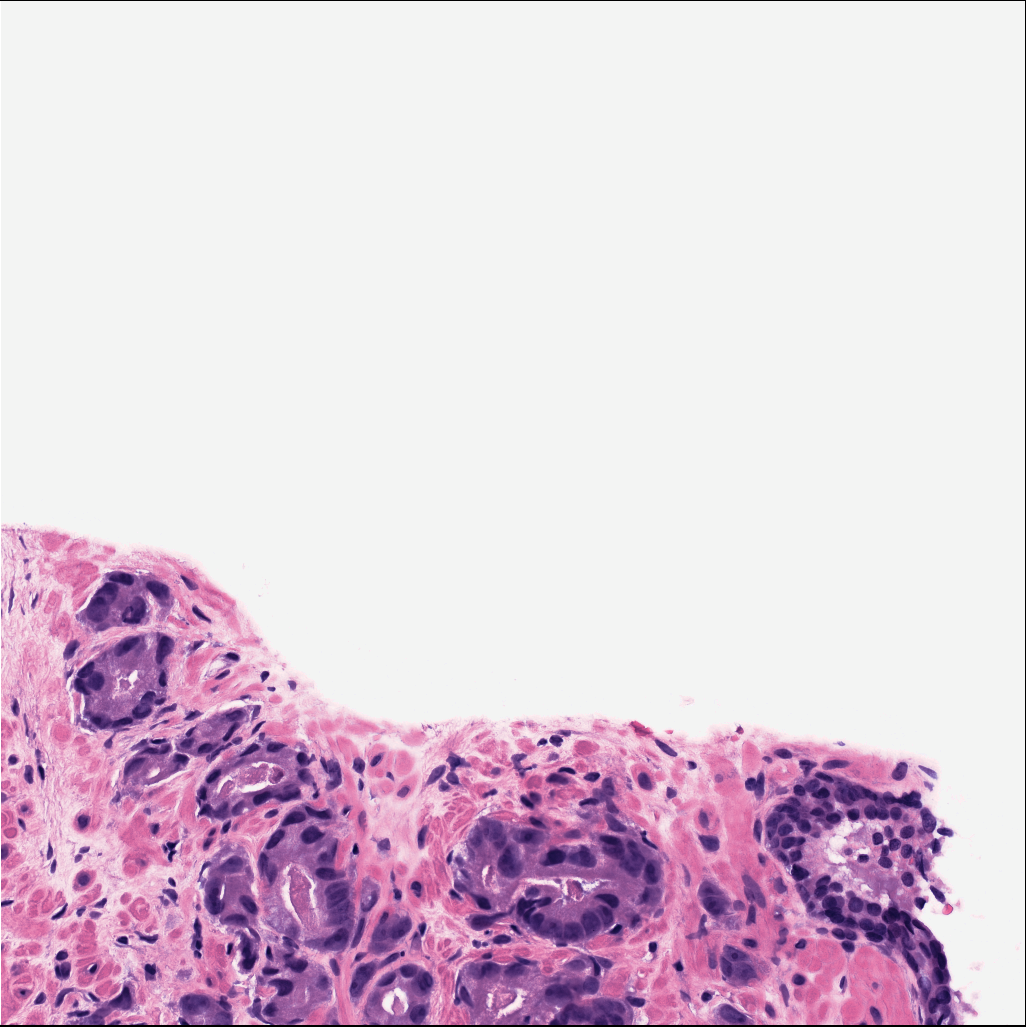} \\

\scriptsize (C) &
\includegraphics[height=2.3cm, align=c]{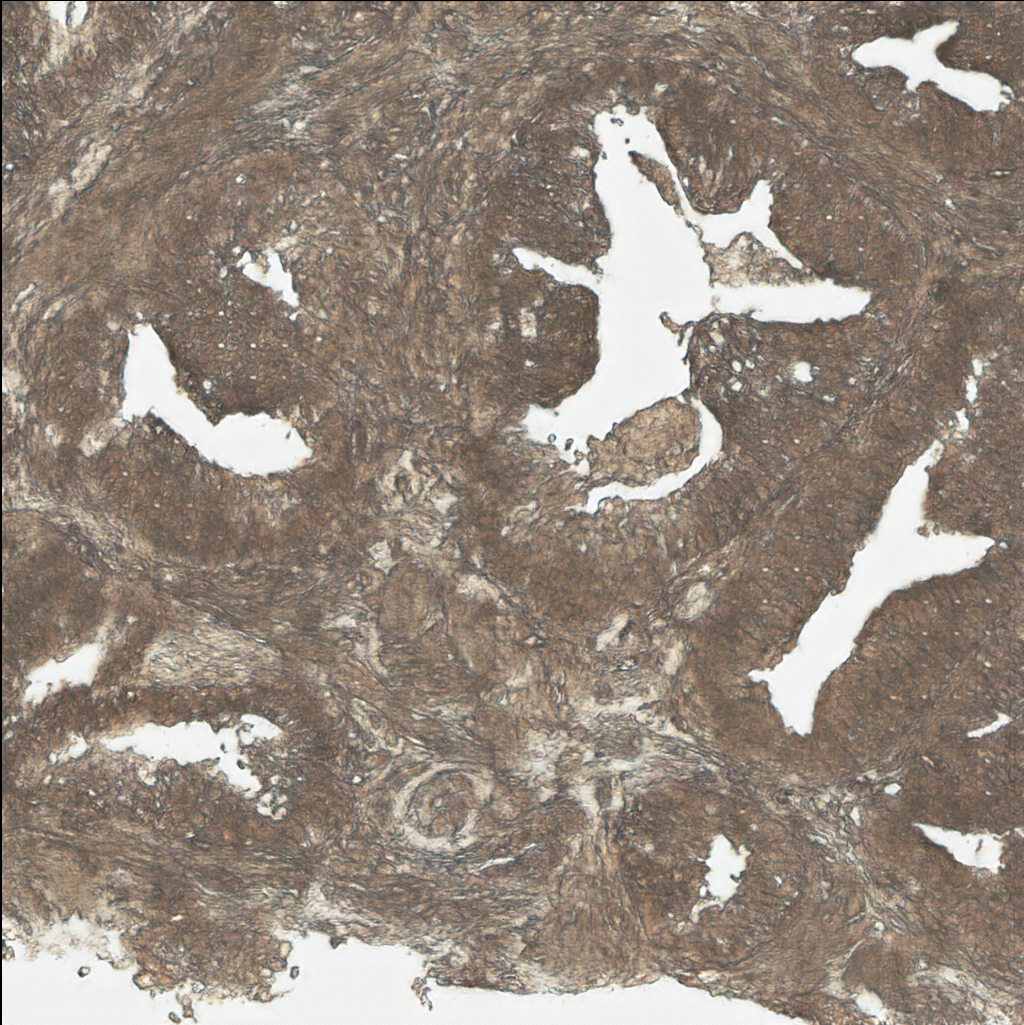} &
\includegraphics[height=2.3cm, align=c]{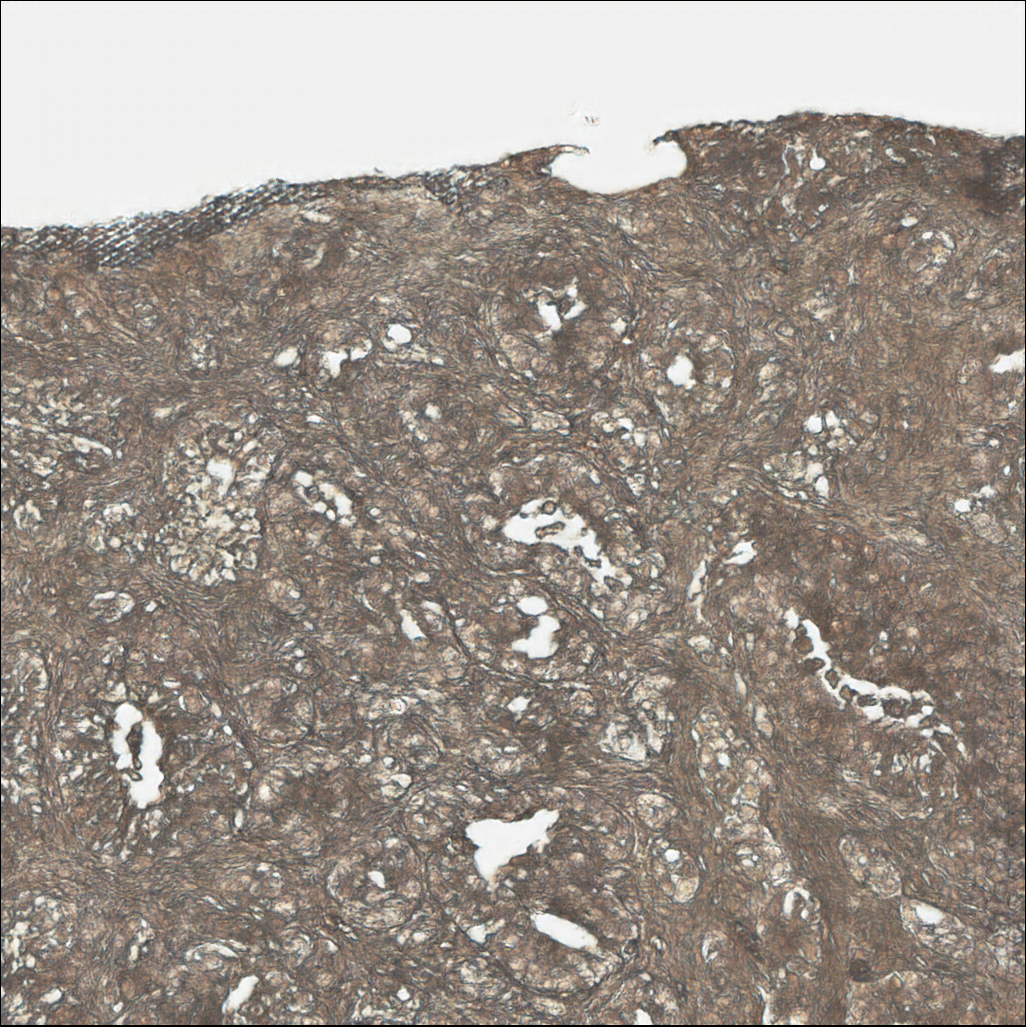} &
\includegraphics[height=2.3cm, align=c]{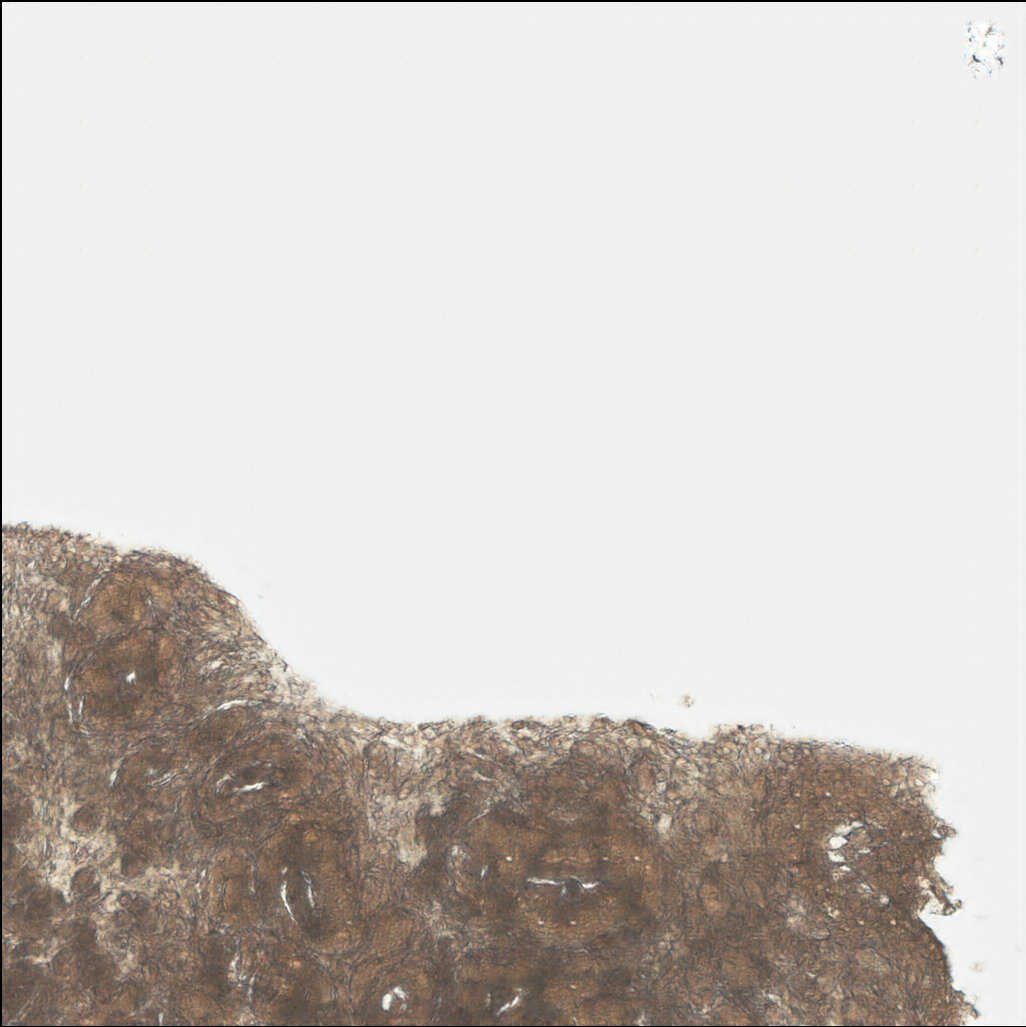} \\

\scriptsize (D) &
\includegraphics[height=2.3cm, align=c]{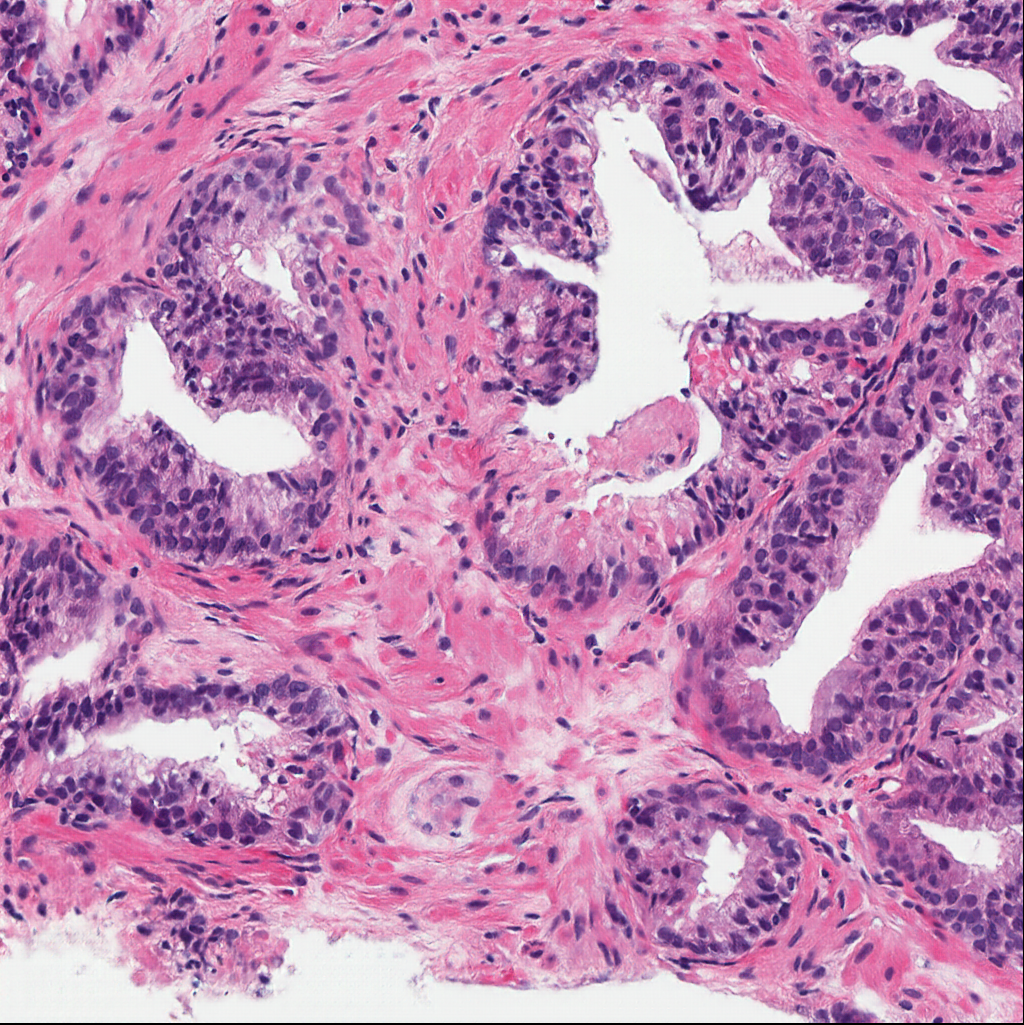} &
\includegraphics[height=2.3cm, align=c]{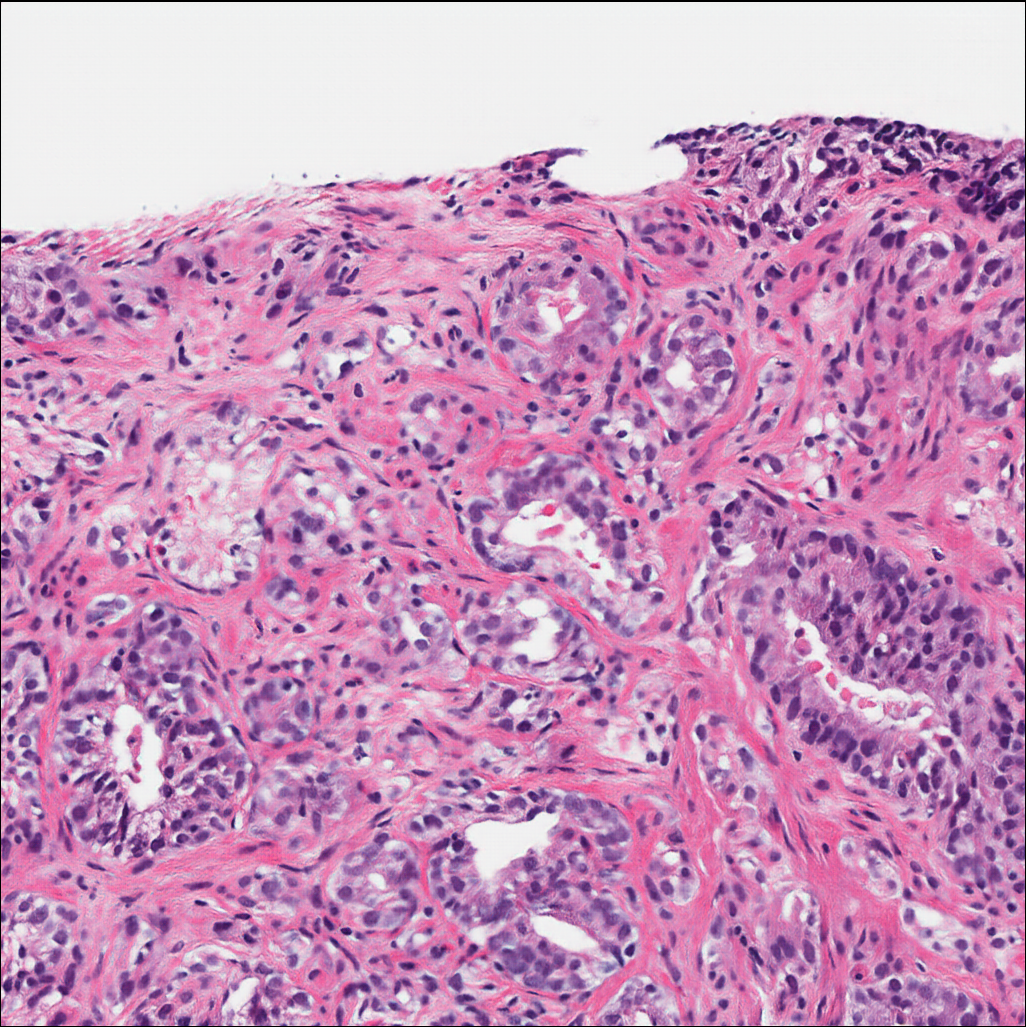} &
\includegraphics[height=2.3cm, align=c]{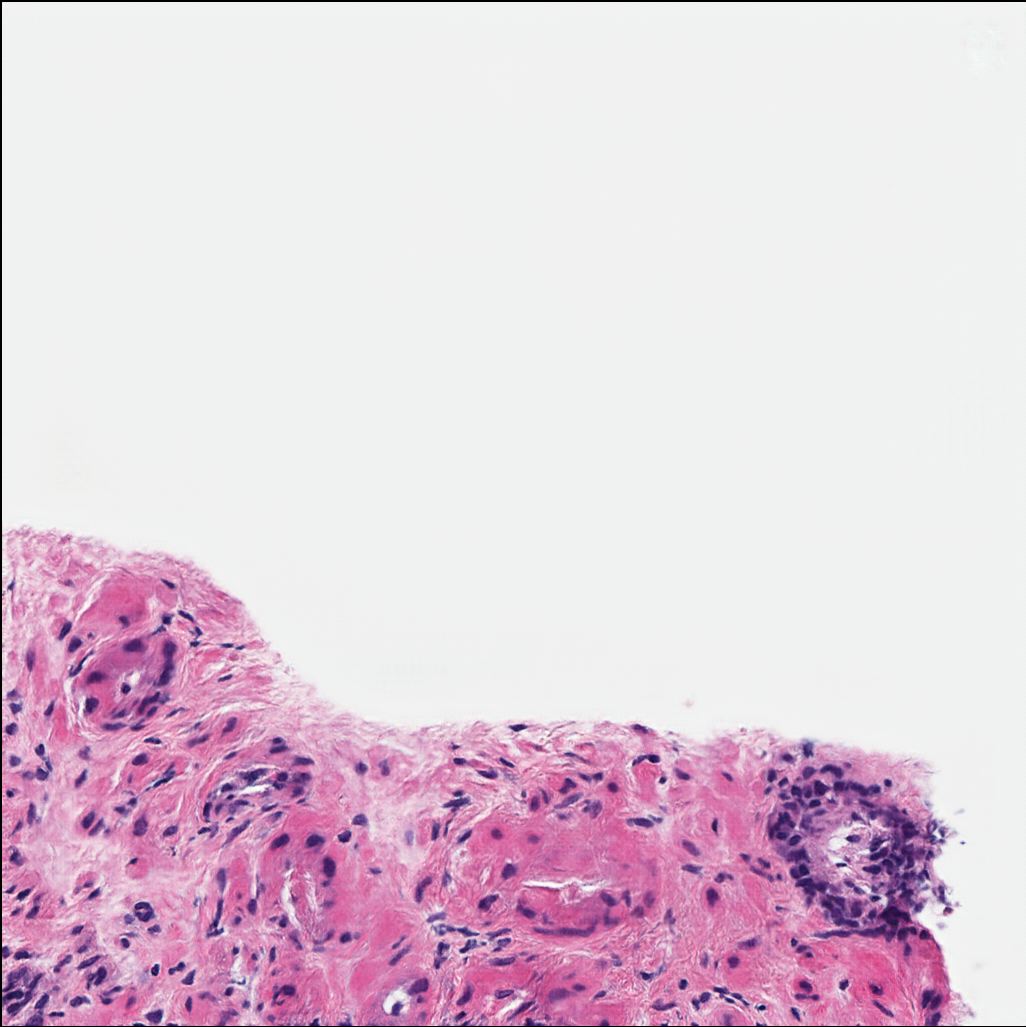} \\

\end{tabular}
\caption{Patch-level examples of ground-truth and generated tissue representations. Three representative patch sets are shown. Row A: ground-truth unstained (GUS). Row B: ground-truth H\&E (GH\&E). Row C: virtually destained (VDS). Row D: virtually H\&E-restained (VH\&ER). According to pathologist review, Column I depicts benign tissue that was well preserved after restaining, Column II depicts malignant tissue with morphology that remained identifiable, and Column III depicts malignant tissue that was rendered as a vessel-like structure.}
\label{fig4}

\end{figure}
Re-inferenced baseline intensity comparisons on the 13 validation WSIs from dataset\_B showed that dataset\_B VDS vs. dataset\_B GUS had a near-zero overall intensity difference of -0.2 ± 0.10 units, whereas dataset\_B VH\&ER vs. dataset\_B GH\&E had an overall intensity difference of -4.60 ± 4.20 units (Table~\ref{tab3}). Relative to these dataset\_B validation baselines, the corresponding dataset\_S comparisons showed substantially larger overall intensity differences, particularly for VDS vs. GUS (-39.8 ± 9.13 in dataset\_S vs. -0.2 ± 0.10 in dataset\_B) and for GH\&E vs. VH\&ER (-6.37 ± 2.82 in dataset\_S vs. -4.60 ± 4.20 in dataset\_B), indicating residual domain shift under external deployment. These findings plausibly help explain both the PCC-SSIM dissociation in the dataset\_S VDS vs. GUS comparison and the stronger performance of the restaining loop relative to direct staining.

A board-certified pathologist reviewed computationally restained images across four prostate biopsy cases representing different cancer burdens and morphologic patterns (Fig.~\ref{fig4}, Rows A--D). Overall, benign tissue compartments were more reliably preserved than malignant glands. Benign glands, stroma, fibroadipose tissue, and broad spatial organization were consistently maintained in the virtual restained images (Fig.~\ref{fig4}, Column I). The pathologist noted that gland-to-stroma-to-fat relationships were generally preserved and that benign glands remained in anatomically plausible locations (Fig.~\ref{fig4}, Column I). The dominant failure mode involved malignant glands. For example, cancerous glands were partially converted into hybrid structures with pink muscular walls resembling blood vessels, replacing the darker purple nuclear morphology expected in malignant glands (Fig.~\ref{fig4}, Column III).

Clinical performance varied by cancer grade and morphology. Case 32, Core 4, which contained predominantly Gleason grade 3 cancer with focal grade 4 features, was the best-performing case. Approximately 80--90\% of malignant gland morphology remained recognizable, and Gleason grade 3 patterns were still identifiable (Fig.~\ref{fig4}, Column II). In contrast, Case 42, Core 10, containing Gleason 3+4 cancer with cribriform and poorly formed patterns, showed loss of diagnosable malignant morphology (Fig.~\ref{fig4}, Column III). An incidental squamous epithelium region in Case 44 provided an additional out-of-distribution test case, since this tissue type was not represented in the dataset\_B training set. The model did not hallucinate inappropriate prostate glandular structures in this region and generated a plausible approximation of the tissue. And the model may preserve broad tissue boundaries in out-of-distribution regions but may not reliably reproduce tissue-specific diagnostic features. At the whole-core level, reconstructed VDS and VH\&ER outputs also preserved the broad spatial layout and tissue organization of the corresponding ground-truth cores despite visible appearance differences. Overall, the computational pipeline preserved benign architecture and broad spatial organization, except the malignant gland morphology and higher-grade and cribriform patterns (Fig.~\ref{fig4}, Columns I--III). 

\section{Discussion and Future Work}
This study reports that under cross-site deployment with preprocessing-based domain adaptation and no retraining, the pretrained cGAN models retained meaningful performance on unregistered dataset\_S WSIs. Virtual destaining achieved the highest PCC among comparison groups, and the destain-restain loop produced H\&E outputs more similar to ground-truth than direct staining. A notable finding was that restaining through the destaining-staining loop outperformed direct computational staining across all reported pixel-level metrics, despite the restained output being generated through two sequential models whereas direct staining used only the staining model. One possible explanation is that the virtually destained intermediate was more closely aligned with the staining model’s learned input distribution than the normalized unstained WSIs from dataset\_S. In that case, the destaining model may have acted not only as a transformation stage but also as an implicit harmonization step that partially reduced mismatch between dataset\_S WSI and dataset\_B training domain. Although this interpretation does not establish causality, it suggests that input harmonization and preprocessing quality may be highly influential in external deployment settings.

Another quantitative observation was the dissociation between PCC and SSIM in the destaining comparison. Virtual destaining against normalized ground-truth unstained tissue WSI achieved the highest PCC but the lowest SSIM among the principal evaluation groups. This suggests that the model preserved strong global pixel correspondence while remaining less consistent in local luminance, contrast, and structural similarity. The intensity analysis supports this interpretation. The GH\&E-to-VDS transformation introduced only a minimal overall intensity shift, but it redistributed RGB channel intensities in a structured manner consistent with reduction of eosin-associated red/pink and hematoxylin-associated blue/purple signal. At the same time, the large uniform intensity gap between VDS and normalized GUS indicates that the unstained-domain preprocessing pipeline likely introduced a variance in brightness and local contrast that SSIM penalized more strongly than PCC. Thus, the destaining appeared to function primarily as a color transformation rather than as a large overall intensity transformation.

The baseline intensity analysis further supports this finding. Re-inferenced baseline comparisons on the 13 validation WSIs from dataset\_B showed that dataset\_B VDS vs. Dataset\_B GUS had a near-zero overall intensity difference, whereas dataset\_B VH\&ER vs. dataset\_B GH\&E retained a modest overall intensity difference. Relative to these dataset\_B validation baselines, the corresponding dataset\_S comparisons showed larger intensity differences under external deployment. After preprocessing, the H\&E normalization pipeline, which combined stain normalization and histogram alignment, likely produced inputs better matched to the dataset\_B training distribution than the heuristic channel-wise calibration used for unstained WSIs. As a result, the staining model may have received a more in-distribution input from the destaining model than from the normalized unstained reference itself. This pattern helps explain two connected findings: first, why VDS vs. GUS retained strong global correlation but lower local structural agreement; and second, why the restaining pathway outperformed direct staining. These results also highlight the role of acquisition and preprocessing variability in digital pathology \cite{niazi2019digital} \cite{flex2025knowledge}. Differences in scanner characteristics, stain preparation, autostainer behavior, section handling, and imaging conditions between dataset\_B and dataset\_S likely contributed to the observed domain shift.

Pathologist review showed that quantitative image similarity did not fully capture diagnostic fidelity. Benign glands, stroma, fibroadipose tissue, and broad spatial relationships were generally preserved in the virtually restained WSIs, whereas malignant glands were more vulnerable to systematic alteration. Case-level variability further suggested that lower-grade cases retained more recognizable malignant gland morphology, whereas higher-grade or more complex patterns, including cribriform and poorly formed glands, were more likely to degrade. Together, these results suggest that the current models preserve broad tissue layout more robustly than fine-grained malignant morphology.

A strength of this study is that it evaluates pretrained models on newly collected, unregistered WSIs after normalization toward the training domain, rather than only on fully paired and spatially registered images from the original development set, making the evaluation closer to real-world deployment \cite{latonen2024virtual, rana2020computational}. At the same time, the reported errors reflect a combination of model error, residual domain shift, and incomplete spatial alignment, and should be interpreted as conservative lower bounds rather than direct equivalents of fully registered single-domain benchmarks. Limitations include a small external cohort from a single outside institution, absence of retraining or alternative domain adaptation baselines \cite{flex2025knowledge, ma2026misalignment}, and specificity to prostate core biopsy WSIs and the pix2pix-based cGAN framework \cite{isola2017image}.

Future work should prioritize stronger unstained-domain harmonization methods, fine-tuning or domain adaptation on target-domain data \cite{flex2025knowledge}, and morphology-aware training objectives that explicitly penalize distortion of malignant glands. Larger multi-institution cohorts with broader pathologist review and direct comparisons between preprocessing-only deployment, fine-tuned models, and registration-aware alternatives will be important for establishing clinical deployment standards. In summary, this study demonstrates that previously trained cGAN models for computational H\&E staining and destaining can retain meaningful performance when applied to newly collected, unregistered prostate biopsy WSIs after normalization toward the training-image domain. The models preserved broad tissue architecture and benign morphology more reliably than malignant gland morphology, and the restaining pathway outperformed direct staining, suggesting that preprocessing normalization may be more limiting than model capacity in this setting. These results suggest that future progress will depend not only on improved generative models, but also on standardized image acquisition, preprocessing pipelines, and open-source tools that support reliable cross-site clinical deployment.

\vspace{-1mm}
\section{Data, Code and Model Availability}
All data, code, and models are available at \url{https://github.com/Dr-Pratik-Shah-UCI/gen_AI_destaining}.





\bibliographystyle{IEEEtran}

\bibliography{sample}

\end{document}